%% file: root.tex
\documentclass[journal, 10pt]{IEEEtran} %

\IEEEoverridecommandlockouts                              %
\usepackage{mathtools,amssymb,commath}
\usepackage{amsmath}
\usepackage{graphicx,import}
\usepackage{verbatim}
\usepackage{color}
\usepackage{hyperref}
\hypersetup{
    colorlinks=false,
    citecolor=[rgb]{0,0,0},
    linkcolor=[rgb]{0.1,0.1,0.1},
    urlcolor=[rgb]{0.0235,0.2706,0.6784},
}
\usepackage{microtype}

\usepackage[table]{xcolor}
\usepackage{booktabs}
\usepackage{multirow}

\usepackage{soul}

\newcommand\change[1]{#1}

\newcommand\urlMystica{\url{https://github.com/ami-iit/mystica}}
\newcommand\urlPaper{\url{https://bit.ly/morph2022}}

\DeclareSymbolFont{matha}{OML}{txmi}{m}{it}%
\DeclareMathSymbol{\varv}{\mathord}{matha}{118}

\usepackage{xfrac}

\usepackage{amsthm}

\theoremstyle{definition}%

\theoremstyle{definition}
\newtheorem{lemma}{Lemma}

\theoremstyle{definition}
\newtheorem{proofLemma}{Proof of Lemma}

\theoremstyle{definition}
\newtheorem{assumption}{Assumption}

\theoremstyle{definition}
\newtheorem{proposition}{Proposition}

\theoremstyle{definition}
\newtheorem{definition}{Definition}

\theoremstyle{definition}
\newtheorem{preliminary}{Preliminary}

\theoremstyle{definition}
\newtheorem{corollary}{Corollary}

\theoremstyle{definition}
\newtheorem{proofCorollary}{Proof of Corollary}

\theoremstyle{remark}

\usepackage{subcaption}

\usepackage{cleveref} %
\crefname{appendix}{appendix}{appendices}

\usepackage{cite}

\makeatletter
\def\@citex[#1]#2{\leavevmode
\let\@citea\@empty
\@cite{\@for\@citeb:=#2\do
{\@citea\def\@citea{,\penalty\@m\ }%
\edef\@citeb{\expandafter\@firstofone\@citeb\@empty}%
\if@filesw\immediate\write\@auxout{\string\citation{\@citeb}}\fi
\@ifundefined{b@\@citeb}{\hbox{\reset@font\bfseries ?}%
\G@refundefinedtrue
\@latex@warning
{Citation `\@citeb' on page \thepage \space undefined}}%
{\@cite@ofmt{\csname b@\@citeb\endcsname}}}}{#1}}
\makeatother

\usepackage{comment}

\usepackage[skins]{tcolorbox}
\usepackage{balance}
\newtcolorbox{myframe}[2][]{%
  enhanced,colback=white,colframe=black,coltitle=black,
  sharp corners,boxrule=0.4pt,left=0pt,right=0pt,top=0pt,bottom=0pt,
  fonttitle=\itshape,
  attach boxed title to top left={yshift=-0.3\baselineskip-0.4pt,xshift=2mm},
  boxed title style={tile,size=minimal,left=0.5mm,right=0.5mm,
  colback=white,before upper=\strut},
  title=#2,#1
}
\DeclareCaptionFont{mysize}{\fontsize{8.5}{9.6}\selectfont}
\include{variablesNamingShortcuts}

\begin{document}

\title{
Modeling and Control of Morphing Covers \\ for the Adaptive Morphology of Humanoid Robots}

\author{Fabio Bergonti$^{1,2}$, Gabriele Nava$^{1}$, Luca Fiorio$^{3}$, \change{Giuseppe L'Erario$^{1,2}$} and Daniele Pucci$^{1,2}$%
\thanks{$^{1}$Artificial and Mechanical Intelligence Laboratory, Fondazione Istituto Italiano di Tecnologia, via San Quirico 19D, Genoa, Italy. {Email addresses: \tt\small firstname.surname@iit.it}}
\thanks{$^{2}$School of Computer Science, University of Manchester,
        Manchester M13 9PL, United Kingdom}
\thanks{$^{3}$iCub Tech, Fondazione Istituto Italiano di Tecnologia, via San Quirico 19D, Genoa, Italy. {Email addresses: \tt\small firstname.surname@iit.it}}
        
}

\import{}{0-abstract}

\markboth{IEEE Transactions on Robotics}%
{Bergonti \MakeLowercase{\textit{et al.}}: Modeling and Control of Morphing Covers for the Adaptive Morphology of Humanoid Robots}

\maketitle
\IEEEdisplaynontitleabstractindextext
\IEEEpeerreviewmaketitle

\section*{Multimedia Material}
The code of this project is available at \urlPaper{}.
Additional results can be viewed in this video: \url{https://youtu.be/kMfXb2xqGn4}.

\import{}{1-introduction}
\import{}{2-background}

\import{}{3-modeling}
\import{}{4-actuatorSelection}
\import{}{5-control}

\import{}{6-simulation}
\import{}{7-conclusions}
\import{}{8-appendix}

\addtolength{\textheight}{0cm}     %

\ifCLASSOPTIONcaptionsoff
  \newpage
\fi

\bibliographystyle{IEEEtran}
\bibliography{IEEEabrv,bibliography}

\vspace{15mm}

\begin{IEEEbiography}[{\includegraphics[width=1in,trim={0 0 0 0},clip,keepaspectratio]{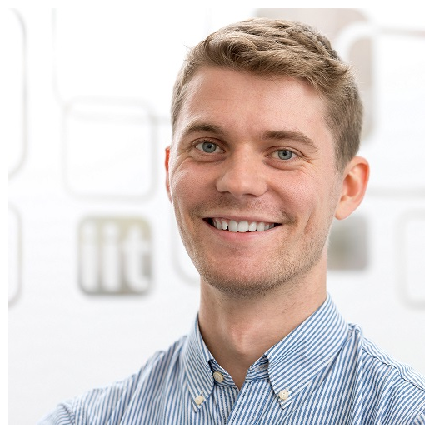}}]{Fabio Bergonti}
received the bachelor’s (Hons.) and master’s (Hons.) degrees in mechanical engineering from the Politecnico di Milano, Milan, Italy, in 2016 and 2018, respectively. He is currently working toward the Ph.D. degree in computer science with the University of Manchester,Manchester, U.K.,working at the Italian Institute of Technology, Genova, Italy, under the supervision of Daniele Pucci and Angelo Cangelosi. He conducted his master thesis at the Dynamic Interaction and Control Laboratory, Italian Institute of Technology, focusing on the development of whole-body control algorithms to generate and control jumps of humanoid robots. From 2018 to 2019, he was a Research Fellow with the Dynamic Interaction and Control Laboratory, Italian Institute of Technology, where he then joined the iRonCub Group, a research group that aims at creating the first flying humanoid robot.
\end{IEEEbiography}

\vspace{-0.5cm}

\begin{IEEEbiography}[{\includegraphics[width=1in,trim={170 0 160 0},clip,keepaspectratio]{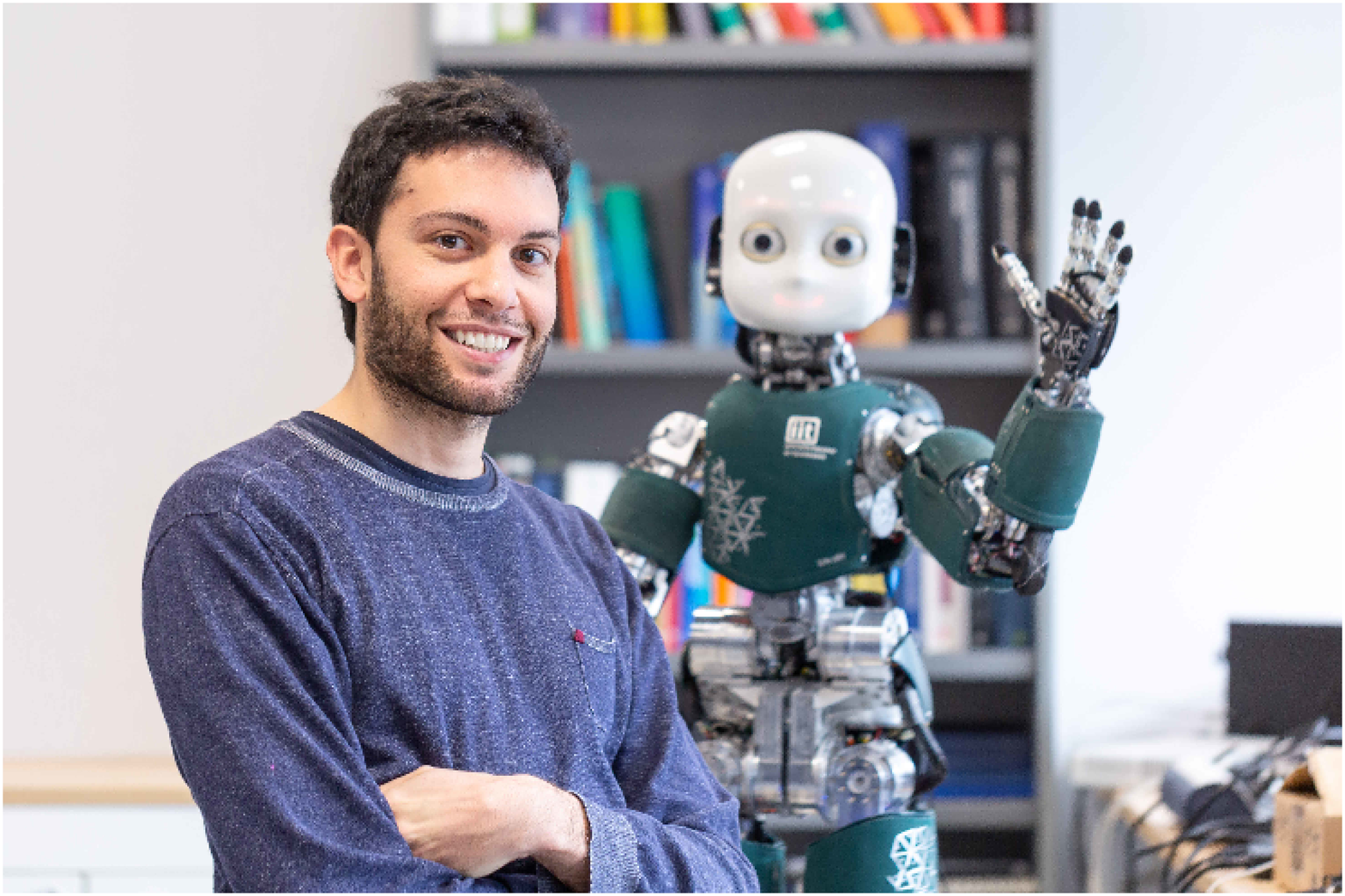}}]{Gabriele Nava}
received the bachelor’s and master’s degrees in mechanical engineering from the Politecnico di Milano, Milan, Italy, in 2013 and 2015, respectively, and the Ph.D. degree in bioengineering and robotics from the Università degli Studi di Genova, Genoa, Italy, in 2020, in cooperation with the Italian Institute of Technology (IIT), Genoa, under the supervision of Daniele Pucci and Giorgio Metta. He is currently a Postdoctoral Researcher with the Artificial and Mechanical Intelligence Laboratory, IIT. His main research interests include the design of control algorithms for floating base systems, with a focus on aerial humanoid robotics. He is the Scrum Master of the iRonCub Group (IIT), which pursues the objective of making the humanoid robot iCub fly.
\end{IEEEbiography}

\vspace{-0.5cm}

\begin{IEEEbiography}[{\includegraphics[width=1in,trim={0 0 0 0},clip,keepaspectratio]{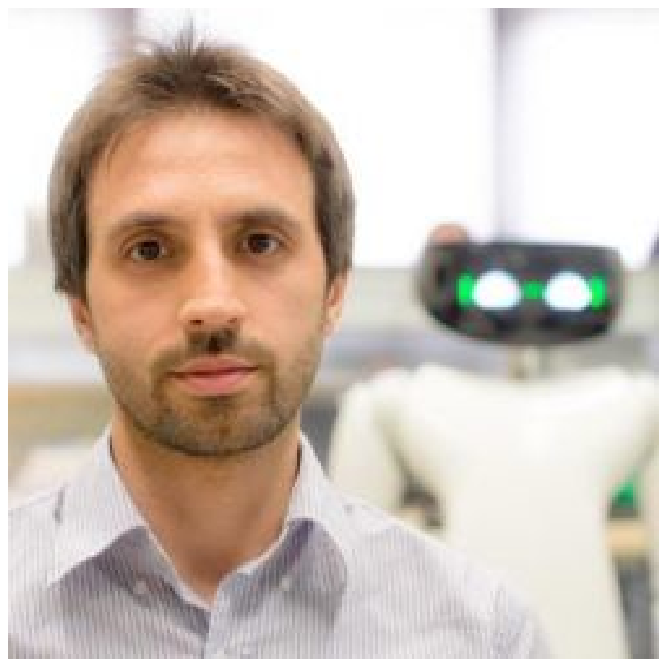}}]{Luca Fiorio}
received the bachelor’s and master’s degrees in mechanical engineering from the Politecnico di Torino, Turin, Italy, in 2007 and 2010, respectively, and the Ph.D. degree in bioengineering and robotics from the Italian Institute of Technology, Genoa, Italy, and the University of Genoa, Genoa, in 2015. He conducted his master thesis with the German Aerospace Center,Munich,Germany, focusing on the simulation and control of a running hexapod robot. Subsequently, as a Postdoctoral Fellow with the Italian Institute of Technology, he worked on the design of the robot R1 and with the Dynamic Interaction Control Laboratory. He is currently a Chief Technician with the iCub Tech Facility, Italian Institute of Technology.
\end{IEEEbiography}

\vspace{-0.5cm}

\begin{IEEEbiography}[{\includegraphics[width=1in,trim={0 0 0 0},clip,keepaspectratio]{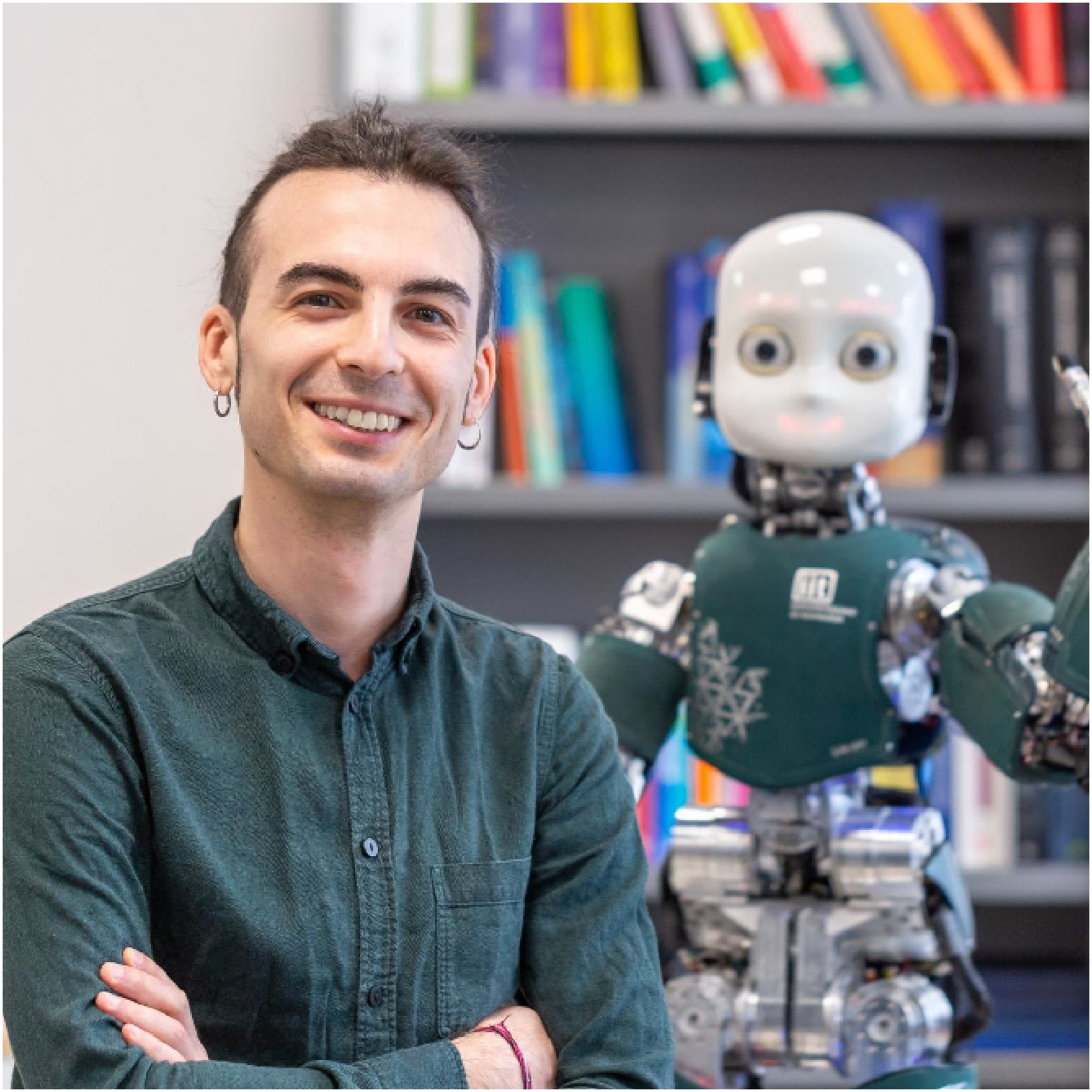}}]{Giuseppe L'Erario}
received the bachelor’s degree in aerospace engineering and master’s degree in artificial intelligence and robotics from the Sapienza University of Rome, Rome, Italy, in 2015 and 2019, respectively. He is currently working toward the Ph.D. degree in computer science with the University of Manchester, Manchester, U.K., working with the iRonCub Group, Italian Institute of Technology, Genoa, Italy, under the supervision of Daniele Pucci and Angelo Cangelosi. He conducted his master thesis with the Dynamic Interaction and Control Lab, IIT, focusing on the modeling, identification, and control of small-scale jet engines, in the context of the iRonCub project, whose aim is to create the first jet-powered flying humanoid robot.
\end{IEEEbiography}

\vspace{-0.5cm}

\begin{IEEEbiography}[{\includegraphics[width=1in,height=1.25in,clip,keepaspectratio]{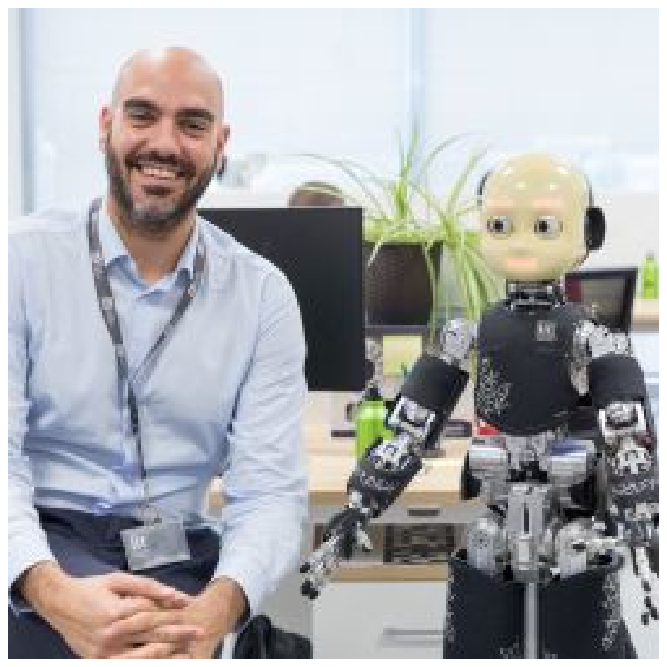}}]{Daniele Pucci}
received the bachelor’s (Hons.) and master’s (Hons.) degrees in control engineering from the Sapienza University of Rome, Rome, Italy, in 2007 and 2009, respectively, and the Ph.D. degree in nonlinear control applied to flight dynamics from INRIA Sophia Antipolis, Biot, France, in 2013, under the supervision of Tarek Hamel and Claude Samson. From 2013 to 2017, he was a Postdoctoral Researcher with the Istituto Italiano di Tecnologia (IIT), Genoa, working within the EU project CoDyCo. Since August 2017, he has been the Head of the Artificial and Mechanical Intelligence Laboratory, IIT. This laboratory focuses on the humanoid robot locomotion problem,with specific attention on the control and planning of the associated nonlinear systems. Also, the laboratory is pioneering aerial humanoid robotics, whose main aim is to make flying humanoid robots. Currently, the laboratory is implementing iRonCub, the jet-powered flying version of the humanoid robot iCub. He is also the scientific principal investigator of the H2020 European Project AnDy, a task leader of the H2020 European Project SoftManBot, and a coordinator of the joint laboratory of IIT and Honda JP. Since 2020, and in the context of the split site Ph.D. supervision program, he has been a visiting Lecturer with the University of Manchester, Manchester, U.K. Dr. Pucci received the “Academic Excellence Award” from the Sapienza University of Rome in 2009. In 2019, he was awarded as Innovator of the Year under 35 Europe from MIT Technology Review.
\end{IEEEbiography}

\vfill

\end{document}

%% file: variablesNamingShortcuts.tex
\newcommand{\brRound}[1]{\left( #1 \right)}
\newcommand{\brSquare}[1]{\left[ #1 \right]}
\newcommand{\brCurly}[1]{\{ #1 \}}

\newcommand{\bm}[1]{\begin{bmatrix} #1 \end{bmatrix}}

\newcommand{\R}[1]{\mathbb{R}^{#1}}
\newcommand{\SO}[1]{\text{SO} \brRound{#1}}
\newcommand{\SE}[1]{\text{SE} \brRound{#1}}

\newcommand{\eye}[1]{{I}_{#1}}

\newcommand{\zero}[1]{\boldsymbol{0}_{#1}}

\newcommand{\tr}{\top}

\newcommand{\skw}[1]{\mathcal{S}{\brRound{#1}} }

\newcommand{\rank}[1]{\text{rank} \brRound{#1} }

\DeclareMathOperator{\dof}{DoF}

\newcommand{\nullSpaceMatrix}[1][]{\mathcal{Z}_{#1} }

\newcommand{\dimV}[1]{\text{dim}{\brRound{#1}} }

\newcommand{\ev}[1]{\boldsymbol{e_{#1}}}

\newcommand{\frameW}{}

\newcommand{\numberSphericalJoints}{N_{\text{sj}}}

\newcommand{\rotm}[2]{R^{#1}_{#2}}
\newcommand{\rotmW}[1]{\rotm{\frameW}{#1}}

\newcommand{\quat}[2]{\mathcal{Q}^{#1}_{#2}}
\newcommand{\quatW}[1]{\quat{\frameW}{#1}}

\newcommand{\dquat}[2]{\dot{\mathcal{Q}}^{#1}_{#2}}
\newcommand{\dquatW}[1]{\dquat{\frameW}{#1}}

\newcommand{\versor}[3]{\boldsymbol{\hat{#1}}^{#2}_{#3}}
\newcommand{\versorW}[2]{\boldsymbol{\hat{#1}}^{\frameW}_{#2}}

\newcommand{\dversorW}[2]{\boldsymbol{\dot{\hat{#1}}}^{\frameW}_{#2}}

\newcommand{\pos}[3]{\boldsymbol{p}^{#1}_{#2,#3}}
\newcommand{\posW}[1]{\boldsymbol{p}_{#1}}

\newcommand{\linVel}[3]{\boldsymbol{\dot{p}}^{#1}_{#2,#3}}
\newcommand{\linVelW}[1]{\boldsymbol{\dot{p}}_{#1}}

\newcommand{\angVel}[3]{\boldsymbol{\omega}^{#1}_{#2,#3}}
\newcommand{\angVelW}[1]{\boldsymbol{\omega}_{#1}}

\newcommand{\node}[2]{\left( #1,#2 \right)}
\newcommand{\nodeij}{\node{i}{j}}
\newcommand{\nodeipj}{\node{i+1}{j}}
\newcommand{\nodeijp}{\node{i}{j+1}}

\newcommand{\J}[1]{J_{#1}}

\newcommand{\Jc}{\J{c}}

\newcommand{\convm}[2]{\mathcal{M}^{#1}_{#2}}

\newcommand{\selm}[1]{S_{#1}}

\newcommand{\permm}[1]{P_{#1}}

\newcommand{\nodesAbsVel}[1][]{\boldsymbol{v}_{#1}}

\newcommand{\nodesRelOmega}[1][]{\boldsymbol{\omega}^{R}_{r_{#1}}}

\newcommand{\nodesRelOmegaAct}{\nodesRelOmega[act]}

\newcommand{\vb}{\boldsymbol{V_\mathcal{B}}}

\newcommand{\VbAndNodesRelOmega}[1][]{\boldsymbol{\nu}_{#1}}

\newcommand{\distJointI}{\boldsymbol{l_{1}}}

\newcommand{\statePosRotm}{\boldsymbol{q}}
\newcommand{\stateVelRotm}{\boldsymbol{\dot{q}}}

\newcommand{\statePosQuat}{\boldsymbol{x}}
\newcommand{\stateVelQuat}{\boldsymbol{\dot{x}}}

\newcommand{\nodePositionError}[1]{\zeta_{#1}}
\newcommand{\normalAlignmentError}[1]{\varepsilon_{#1}}

\newcommand{\wrt}{\text{w.r.t. }}

\newcommand{\ie}{\text{i.e. }}
\newcommand{\eg}{\text{e.g. }}

%% file: 0-abstract.tex
\IEEEtitleabstractindextext{
    \begin{abstract}
        This article takes a step to provide humanoid robots with adaptive morphology abilities. We present a systematic approach for enabling robotic covers to  morph their shape, with an overall size fitting the anthropometric dimensions of a humanoid robot.
        More precisely, we  present a cover concept consisting of two main components: a \emph{skeleton}, which is a repetition of a basic element called \emph{node}, and a soft \emph{membrane},  which encloses the cover and deforms with its motion. This article focuses on the cover skeleton and addresses the challenging problems of node design, system modeling, motor positioning, and control design of the morphing system. The cover modeling  focuses on kinematics, and a systematic approach for defining the system kinematic constraints 
        is presented. Then, we apply genetic algorithms to find the motor locations so that the morphing cover is fully actuated. Finally, we present control algorithms that allow the cover to morph into a time-varying shape.
        The entire approach is validated by performing kinematic simulations with 
        \change{four} different covers of square dimensions and having $3\times3$, \change{$4\times8$}, $8\times8$, and $20\times20$ nodes, respectively. For each cover, we apply the genetic algorithms to choose the motor locations and perform simulations for tracking a desired shape.
        The simulation results show that the presented approach ensures the covers to track a desired shape with \emph{good} tracking performances. %
    \end{abstract}
    
    \begin{IEEEkeywords}
        Adaptive morphology, genetic algorithm, humanoid robots, motion control, parallel robots, robot kinematics.
    \end{IEEEkeywords}
}

%% file: 1-introduction.tex
\section{Introduction}

\IEEEPARstart{A}{daptive} morphology is a fascinating natural feature that Robotics struggles with implementing into the existing platforms. Birds~\cite{modelingFlyingBirdChap5}, snakes~\cite{flyingSnake}, and frogs~\cite{morphingFrog,flyingFrog} are only a few examples of the large variety of animals that use adaptive morphology for several goals, ranging from increasing locomotion abilities to augmenting self-defense systems~\cite{salamander}. Although successful attempts of implementing adaptive morphology into robotic platforms exist~\cite{adaptiveMorphologyFloreano2016,morphingQuadrotorScaramuzza2019,roboticSkinBottiglio2019,shah2020shape}, humanoid robots are still 
conceived with links having rigid covers, which limit their capacity to adapt to the surrounding environment and to optimize locomotion and manipulation tasks. This article proposes a concept of morphing covers for humanoid robots and addresses the associated fundamental problems of kinematic modeling, actuation positioning, and control design of the morphing system.

\begin{figure}[t]
        \begin{subfigure}[b]{0.49\columnwidth}
            \centering
            \includegraphics[width=\textwidth]{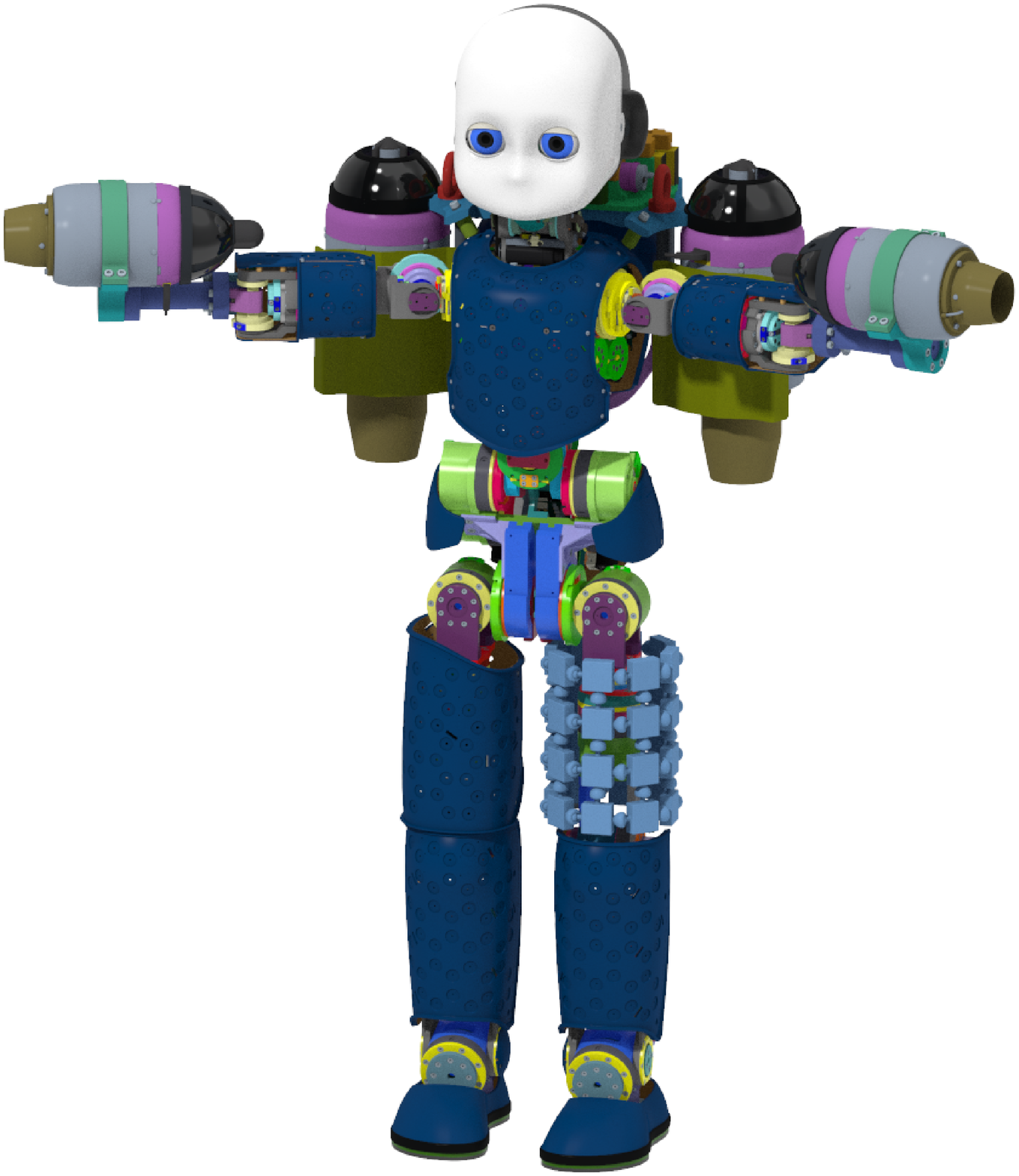}
            \caption[]{Rest configuration}
            \label{fig:iRonCub-rest}
        \end{subfigure}
        \hfill
        \begin{subfigure}[b]{0.49\columnwidth}  
            \centering 
            \includegraphics[width=\textwidth]{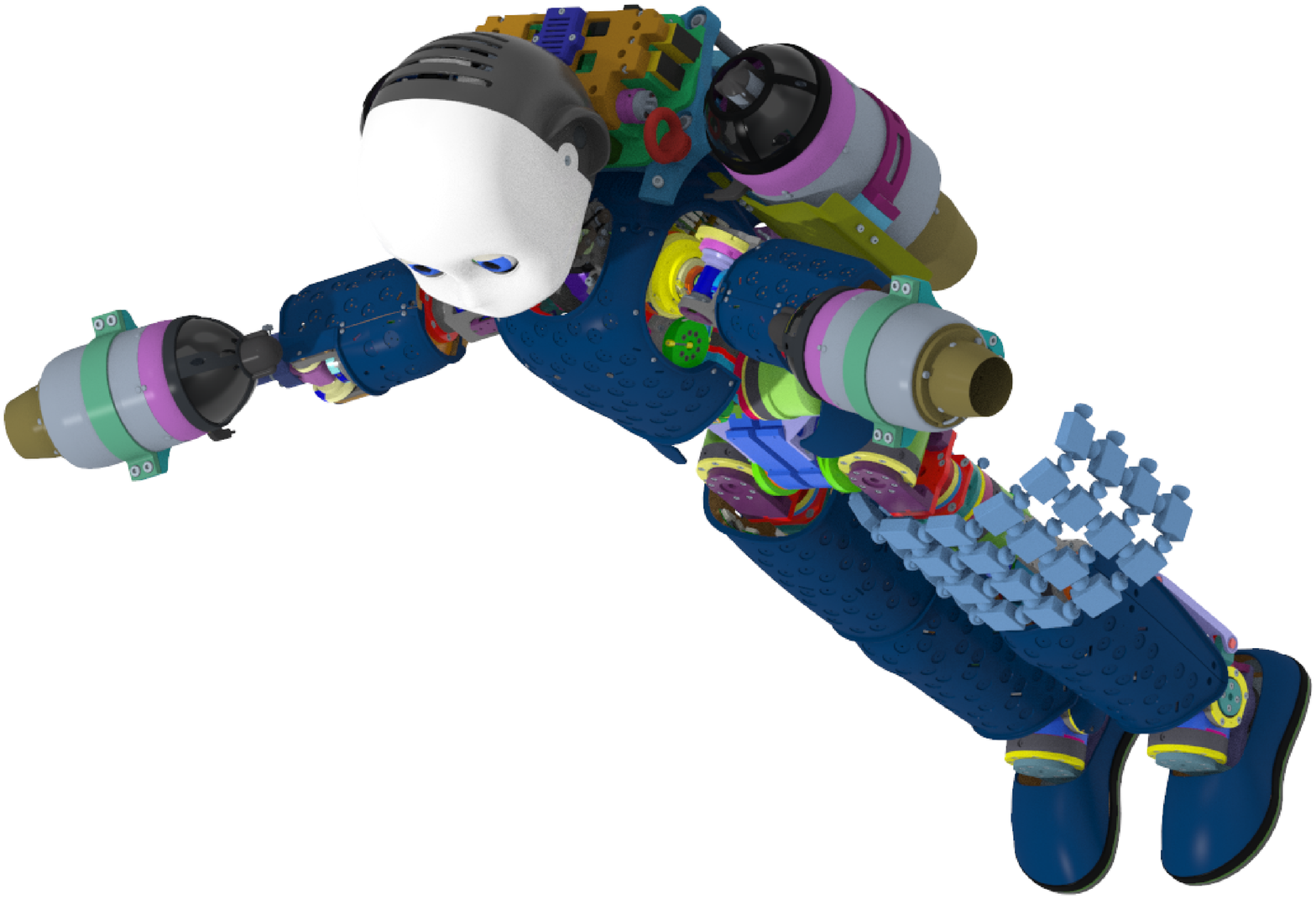}
            \caption[]{Morphed configuration}
            \label{fig:iRonCub-open}
        \end{subfigure}

    \caption[]{\change{Conceptual picture of the \emph{iRonCub}  robot with a morphing cover integrated in the left thigh. The cover morphs and adapts its shape to optimize aerodynamics. The real \emph{iRonCub} is currently being tested to perform the first flight}~\cite{ironcubHosam}.}
    \label{fig:iRonCub}

\end{figure}

Most of humanoid robot  design is  governed by the idea of having a skeleton enclosed by rigid covers \cite{natale2017icub, atlasArticle, asimo2019, surveyHumanoidRobot2019}.
The covers are the first robot body part that makes contact with the 
environment, so they have a pivotal role during any robot \emph{physical} task -- \eg, locomotion, human-robot interaction, manipulation, etc.
Consequently, adaptive morphology via morphing covers may lead to unprecedented features of humanoid robots. Morphing feet that optimize locomotion patterns (\eg flat feet for balancing and point/rounded feet for walking) and adaptive chest shapes to pass through narrow cavities are only a few leaps forward that morphing covers of humanoid robots would enable. 
\change{%
Our long-term vision is to provide future flying humanoid robots  the ability  to morph their shape to optimize the resulting aerodynamics acting on the robot. We are currently implementing the vision of morphing flying humanoid robots on iRonCub}\cite{ironcubRal,ironcubHumanoids}\change{, a modified version of the iCub robot powered by turbojet engines integrated in the chest and in the arms -- see Fig.~}\ref{fig:iRonCub}.

\change{The idea of enhancing robot capabilities via cover and link re-design is not new to the Robotics community.
Examples are: the integration of soft materials as robot skin to perceive contacts and absorb collisions} \cite{humanoidRobotSoftSensors,hakozaki1999telemetric,humanoidRobotShockAbs}\change{, or the design of links with an active variable stiffness control~}\cite{morphingRoboticArm2017}.
\change{However, how to  implement effectively morphing abilities of humanoid robots is still an open issue although there is an active scientific effort in tackling several problems around the topic -- see, e.g., the problems of shape morphing structures, surface control, and transformation from 2D surfaces into 3D bodies. We briefly review below the main approaches for tackling these problems.}

\change{The ancient art of \emph{origami} gave impetus to researchers in proposing folding mechanisms capable of evolving from $2$-D to $3$-D structures~}\cite{origamiLang1996,rus2018origamiRobotReview}. \change{At the actuation level, the folding feature may be driven by pneumatic actuators} \cite{metaOrigamiPneumoActuated2016,softMorphingOrigami2019}\change{, thread-based mechanisms powered by motors }\cite{origamiBot2014}\change{, smart memory alloys }\cite{origamiActuatedRuss2014}\change{, and heat sources }\cite{walkingOrigamiRus2015}. 
\change{\emph{Origami} multi degree-of-freedom ($\dof$) mechanisms can also play the role of actuation mechanisms for \emph{reconfigurable surfaces}} \cite{recSurfOriPixelEPFL2020}.
\change{%
In this case, the resulting \emph{reconfigurable surface} much differs from the classical configuration of having
pixels that can  translate only~}\cite{recSurfMouldingPatent,recSurfESA,recSurfFesto, recSurfInForm2013}.
In general, \change{\emph{reconfigurable surfaces} are usually characterized by an actuation system that generates absolute movements of the pixels and the actuator's stators are fixed to ground. Therefore, reconfigurable surfaces little fit the application of morphing covers for humanoid robots.}

\change{\emph{Soft robots} may implement morphing features since they are 
often made of hyperelastic materials with Young's modulus in the order of $10^4$ and $10^9$}\cite{softRobotReview2015,softRobotReview2017,softRobotReview2018}.
\change{Examples of soft robots that implement a degree of controllable morphing feature are active textiles with McKibben muscles} \cite{activeTextile2017,activeTextile2019}\change{, silicon-rubber pad} \cite{softRoboticPad2017}\change{, buckling surfaces }\cite{bucklingSoftGel2012}\change{, groove-based mechanism }\cite{morphingPasta2021}\change{, heat-sealing inflatable materials pneumatically activated }\cite{aeroMorph2016}\change{, surfaces textures for camouflage} \cite{camouflageRobot2013,camouflageRobotScience2017,pneUISoftRobot2013}\change{, and  underwater morphing robots to optimize hydrodynamic forces} \cite{underwaterWalkingRobot2019}.
\change{Most of these, however, are designed to achieve a unique or a limited range of configurations. A proof of concept to enable a larger spectrum of controllable motions is a morphing surface actuated with liquid crystal elastomers }\cite{softElastometerDaraio2021}.

\change{%
\emph{4D printed variable shaping structures} leverage a new frontier of manufacturing, the $4$-D printing~}\cite{4Dprinting2019}.
\change{\emph{4D printing} consists of $3$-D printed structures that evolve with time when exposed to a predetermined stimulus such as water }\cite{4DprintingWater2014}\change{, temperature }\cite{4DprintingTemperature2019}\change{, magnetic fields}\cite{magneticSoftRobot2018}\change{, etc.}

\change{\emph{Morphing wings} are innovative  flight supports capable of changing their shape and adapt to different aerodynamics conditions }\cite{morphingWingReview2010,morphingWingReview2011}\change{. The challenges when designing \emph{morphing wings} are numerous, ranging from mechanism design to the development of deformable skins that shall withstand the aerodynamic loads}\cite{morphingSkins2008}.
\change{\emph{Morphing wings} can, thus, reconfigure the planform~}\cite{planformMorphingWing2004,planformMorphingWing2007,optimalActuatorMorphingWing2006}\change{, perform out-of-plane transformation}\cite{oopMorphingWing2007}\change{, and adjust the airfoil profile}\cite{airfoilMorphingWing2013}.

\emph{Self-reconfigurable robots}
change their shape by rearranging the connectivity of their constitutive parts~\cite{modularRobotReview2007,modularRobotReview2015,morpho2008}. For instance, modular self-folding robots can be composed of 
a repetition of triangular modules~\cite{mori2017}. From an actuation standpoint, these modules may be either active or passive, but they are both given with the capability of coupling via a specific actuation mechanism~\cite{moriAutomaticCoupling2019}.
The main characteristic of these self-reconfigurable robots is their \emph{low} thickness that allows the mechanism to fold. Consequently, one can generate $3$-D origami from an initial $2$-D layout.
On one hand, the  concept of modular robots  leads to a high structural versatility since the repetition of a basic module allows us to obtain 
\emph{complex}  shapes. Furthermore, the robot design pivots around the basic element to repeat and not around the entire mechanism. On the other hand, self-reconfigurable robots are designed to rearrange their connectivity, so they are not ideal in the case a single connectivity pattern should be kept by the overall platform.

\change{The survey above highlights some of the existing technologies that may be considered when attempting to implement a degree of morphing feature onto a humanoid robot.
Our specific use-case of \emph{morphing covers} for (flying) humanoid robots, however,
introduces dimensional and structural constraints that call for a word of caution. 
From the mechanical standpoint, the cover material shall be stiff enough to resist the aerodynamic pressure without deforming or breaking apart, while  
the  cover 
size shall ease its integration in a human-sized robot. 
From the functional standpoint, the morphing cover shall have a degree of maneuverability  to induce enough variations and benefits in terms of robot shape and aerodynamic loads. Moreover, the cover shall be robust against air pressure and avoid (at most) hysteresis behaviors. The range of achievable shapes shall be various and not limited to bistable states only.
Finally, the cover actuation shall be easily reproducible and of simple integration into the mechanics of the humanoid robot.}

\change{According to these use-case constraints, we believe that none of the existing methods can be applied \emph{as such} for the design of morphing covers for humanoid robots. Rather, we believe that the morphing cover design and control shall combine ideas and principles developed by the \emph{soft} and \emph{self-reconfigurable} robot literature. We, thus, propose a morphing cover composed of  a \emph{skeleton}, namely a rigid mechanism that enables the cover to move, and a \emph{soft external membrane}, namely an elastic material that encloses the cover to exploit aerodynamic forces by filling cavities and passively deforms with its motion. 
Although the concept of \emph{shape morphing} is usually associated with soft structures, 
we refer to as \emph{morphing cover} the proposed rigid skeleton and soft membrane 
because the system macroscopically morphs from one shape to another; 
once integrated into a  humanoid robot, the covers evolution cause a morphological variation of the robot.
The article, however, focuses on the cover skeleton only, and the analysis of the soft membrane will be the subject of future studies.
The cover skeleton 
is composed of a repetition of a rigid basic element, called \emph{node}. The node is of square shape and is connected to the four adjacent nodes via spherical joints. Joints can be actuated, and the problem of finding which joint to actuate is also investigated.
Differently from the \emph{self-reconfigurable} literature, our approach is  optimized for the case where the connectivity pattern of all nodes does not change.}

\change{At the theoretical level, 
this article presents a systematic, model-based, and scalable approach for the modeling, actuation positioning, and control of a morphing system.
The system modeling is approached using maximal coordinates, which allows us to easily tackle kinematics with closed loops. The modeling approach differs from the state of the art  related to similar self-reconfigurable robots that use minimal coordinates} ~\cite{moriReconfigurationStrategy2019}. %
\change{Motor positioning is tackled using genetic algorithms, thus defining a minimal set of actuators that enable the design of global stabilizers of 
desired $3$-D shapes. 
The proposed motor positioning and control design much differ from the existing methods applied to morphing systems that mostly deal with:}
\change{
$i)$ strategies to place active modules optimized for open chain mechanism} \cite{moriOptimalDistributionActiveModules}\change{; 
$ii)$ definitions of optimal $2$-D layout to obtain a $3$-D shape relying on origami theory}\cite{moriReconfigurationStrategy2019}\change{; 
$iii)$ planning algorithms to define the sequence of foldings}\cite{moriReconfigurationStrategy2019}\change{
.}

More precisely and compactly, 
the contributions of this article are as follows:
\begin{enumerate}
    \item the design of the morphing cover skeleton and node \change{with the aim of applying it to a humanoid robot}; 
    \item the development of the cover kinematic model;
    \item a genetic algorithm for selecting the motor locations so as to obtain a fully actuated cover;
    \item the control design for the morphing cover allowing it to track a time-varying desired $3$-D shape;
    \item the validation with simulation results verifying the robustness of the approach against noise.
\end{enumerate}

Although the contributions above can also be  applied to other natures of robots, we are currently investigating the use of a specific kind of motor that induces a  node dimension in the order of $25$ [mm]. In turn, the node size implies an  overall cover dimension that fits the anthropometric dimensions of existing humanoid robot covers -- see section \ref{sec:simulations}.

The rest of this article is organized as follows. 
Section~\ref{sec:background} introduces notation and recalls the %
basis of genetic algorithms.
Section~\ref{sec:modeling} presents the kinematic model of the proposed morphing cover.
Section~\ref{sec:actuatorChoice} describes a genetic algorithm for motor optimal placing.
Section~\ref{sec:control} presents a framework for the control design of the morphing cover.
Section~\ref{sec:simulations} shows the simulations results.
Finally, Section~\ref{sec:conclusions} concludes this article.

%% file: 2-background.tex
\section{Background} \label{sec:background}

\subsection{Notation}

\begin{itemize}
    \item \change{$\SO{3} := \brCurly{ \rotm{}{} \in \R{3 \times 3} \; | \; \rotm{\tr}{} \rotm{}{} = \eye{3}, \, \det \brRound{\rotm{}{}} = 1}$}
    \item \change{$\SE{3} := \brCurly{ \begin{bsmallmatrix} \rotm{}{} & \posW{} \\ \zero{} & 1 \end{bsmallmatrix} \in \R{4 \times 4} \; | \; \rotm{}{} \in \SO{3}, \, \posW{} \in \R{3}}$  }
    \item $\skw{\boldsymbol{x}}  \in \R{3 \times 3}$ is the skew symmetric matrix associated with the cross product in 
    $\R{3}$, namely, given $\boldsymbol{x},  \boldsymbol{y} \in \R{3}$, then $\boldsymbol{x} \times \boldsymbol{y} = \skw{\boldsymbol{x}} \boldsymbol{y}$.
    \item The vectors $\ev{1}$, $\ev{2}$, $\ev{3}$ denote the canonical basis of $\R{3}$  
    \item $w$ denotes the inertial (or world) frame; $a$, $b$, $c$ are generic body-fixed frames.
    \item $\posW{a} \in \R{3}$ denotes the origin of the $a$ frame expressed in the inertial frame $w$.
    \item $\rotm{a}{b} \in \SO{3}$ is the rotation matrix that transforms a $3$-D vector expressed with the orientation of the $b$ frame in a $3$-D vector expressed in the $a$ frame, \ie, %
    \begin{equation*}
        \rotm{a}{b} = \bm{ \versor{x}{a}{b} & \versor{y}{a}{b} & \versor{z}{a}{b}} \quad \text{with} \quad \versor{x}{a}{b} = \rotm{a}{b} \ev{1}
    \end{equation*}
    $\versor{x}{a}{b}$, $\versor{y}{a}{b}$, and $\versor{z}{a}{b}$ are the unit vectors of the frame axes. %
    \item $\quat{a}{b} \in \R{4}$ denotes the unit quaternion representation of the rotation matrix $\rotm{a}{b}$.    
    \item $\pos{a}{b}{c} \in \R{3}$ is the relative position of $c$ frame \wrt to $b$ frame written in $a$ frame, \ie
    $\pos{a}{b}{c} = \rotm{a}{w} \brRound{ \posW{c} - \posW{b} }$. %
    \item $\linVel{a}{b}{c} \in \R{3}$ is the linear velocity of $c$ frame \wrt to $b$ frame written in $a$ frame.
    \item $\angVel{a}{b}{c} \in \R{3}$ is the angular velocity of $c$ frame \wrt to $b$ frame written in $a$ frame.
    \item $\vb \in \R{6}$ is the $6$-D velocity of base link. %
\end{itemize}

\noindent
Given the vectors $\pos{a}{b}{c}$, $\linVel{a}{b}{c}$, $\angVel{a}{b}{c}$, and $\quat{a}{b}$ or the matrix $\rotm{a}{b}$, if one of the terms 
$a$, $b$, or $c$ is omitted, it is implicitly assumed to be equal to the inertial frame $w$.

\subsection{A Short Recap of Genetic Algorithms}
Genetic algorithms (GA) are search algorithms inspired by the process of natural selection \cite{goldberg1988genetic}. 
Given an initial population composed of a set of candidates, an \emph{evolution process} improves iteratively the characteristics of the populations according to a metric defined by a \emph{fitness function}.
The fitness function is the objective function that we want to maximize (or minimize).
Often, a population member is a scalar number or an array encoded into binary strings.

The evolution of the population is performed using three operators \cite{mitchell1998introduction}: %
    $i)$ reproduction;
    $ii)$ crossover;
    $iii)$ mutation.
\emph{Reproduction} is the selection process of some candidates from the current population. The selection methods are many, \eg,  \emph{fitness  proportionate}  defines the probability of selecting a candidate to be proportional to its fitness~\cite{selectionGA}. %
\emph{Crossover} is a stochastic method to generate new population members from an existing one. \emph{Single-point crossover} is a common implementation, where the strings of two candidates are swapped at a specific string location~\cite{crossoverGA}. %
\emph{Mutation} is a process that alters randomly a population member. It enables the algorithms to look for new solutions and to avoid getting stuck into local minima having suboptimal fitness values.

%% file: 3-modeling.tex
\section{Modeling the Morphing Cover Skeleton} \label{sec:modeling}

    \change{This section presents the skeleton of the morphing cover and the associated kinematic modeling. 
    }

\subsection{Cover Skeleton and Mathematical Preliminaries}

\begin{figure}[t]
            \centering 
            \includegraphics[trim=65 48 55 48, clip, width=0.65\columnwidth]{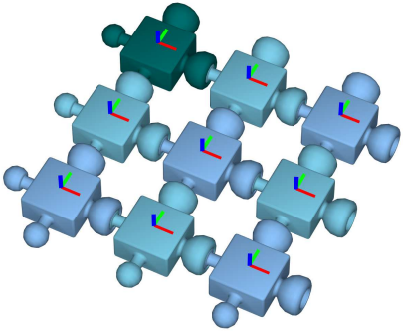}
        
        \caption[]{$3\times3$ morphing cover skeleton. The coordinate frames follow RGB convention. \change{The green-colored node is fixed to the ground.}}
        \label{fig:3by3}
\end{figure}

The proposed morphing cover is composed of two main components: a \emph{cover skeleton}, \ie, a rigid mechanism that provides the cover with the ability to move, and a \emph{soft membrane}, \ie, an elastic material that encloses the cover and that passively deforms with its motion.
As mentioned in the introduction, this article focuses on the skeleton design, modeling, and control. In particular, the skeleton is a mechanism composed of a repetition of a rigid element, called \emph{node}. Here, we propose a node of square section that is connected to the adjacent nodes via spherical joints.
For example, Fig.\ref{fig:3by3} shows a morphing cover skeleton composed of $9$ nodes and $12$ joints. 
In the reminder, we will often refer to the morphing cover skeleton as  \emph{mesh}, or simply as \emph{morphing cover}.

\begin{figure}[t]
    \centering
    \includegraphics[width=\columnwidth]{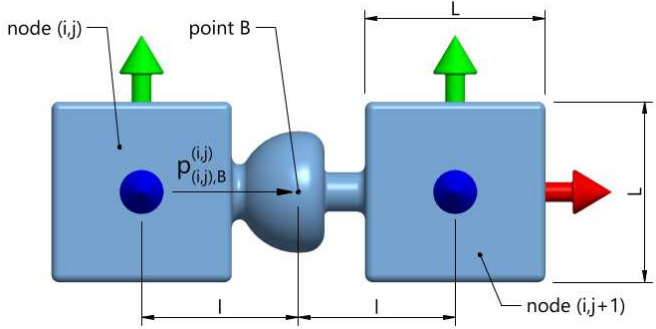}
    \caption[]{Connection between node $\nodeij$ and $\nodeijp$. 
    The point $B$ is the center of the spherical joint. %
    }
    \label{fig:sphericalJoint2DCreo}
\end{figure}

The following presents  mathematical preliminaries, definitions, and assumptions that are  used in the rest of this article. %

\begin{assumption}
    The morphing cover skeleton is composed of $nm$ nodes, with $n$ and $m$ being two natural numbers. 
\end{assumption}

\begin{preliminary}
    Each node is uniquely identified with a pair of natural numbers $\nodeij$, with $i \in \brCurly{ 1,2, \cdots, n }$ and $j~\in~\brCurly{ 1,2, \cdots, m }$. The numbers $\nodeij$ are called \emph{row} and \emph{column} indexes, respectively.
\end{preliminary}

\begin{definition}
    The node of indexes $\nodeij = \node{1}{1}$ is called \emph{father node}, or \emph{base link}. Being of square section, two orthogonal directions can be associated with the father node, here referred to as \emph{row direction} and \emph{column direction}. 
\end{definition}

\noindent For example, the column and row directions may be represented by the red and green arrows in Fig.\ref{fig:sphericalJoint2DCreo}, respectively. Now, we can define how the morphing cover skeleton is built. 

\begin{preliminary} \label{preliminary:built-skeleton}
    Arrange the nodes so that all their square sections belong to the same plane. Consider the \emph{father node} $\node{1}{1}$ and its row and column directions.
    We, then, put the node of indexes $\node{2}{1}$ along the row direction at a distance $2l$ from node $\node{1}{1}$, and we connect the two aforesaid nodes via a spherical joint.
    Each joint has three degrees of freedom ($\dof$). We repeat the operations by connecting in series the nodes of indexes $\node{3}{1}$, $\node{4}{1}$, $\cdots$, $\node{n}{1}$ via spherical joints. Then, we go back to the \emph{father node}, and we place the node of indexes $\node{1}{2}$ 
    along the father node column direction. Again, we connect the nodes via a spherical joint. Then, using spherical joints, we connect in series the nodes $\node{2}{2}$, $\node{3}{2}$, $\cdots$, $\node{n}{2}$ along the row direction. This time, however, each node is also connected by columns, \eg, the nodes $\node{2}{2}$ and $\node{2}{1}$ are also connected via spherical joints. We repeat the overall procedure by incrementing the row and column indexes accordingly. We stop when all $nm$ nodes are displaced in a \emph{matrix-like} mesh.
\end{preliminary}

At the end of this procedure, we obtain a cover skeleton composed of $nm$ nodes displaced in a matrix-like mesh: all the nodes are connected by spherical joints -- see, \eg, Fig.\ref{fig:3by3}. Observe that the total number $\numberSphericalJoints$ of spherical joints in the cover is given by $\numberSphericalJoints = 2nm-n-m$. Finally, we place reference frames in each node composing the morphing cover.
\begin{preliminary}
    Each node is associated with a coordinate frame placed at the center of the square section. More precisely, a node of indexes $\nodeij$ is associated with the pair $\brRound{ \posW{\nodeij},\rotmW{\nodeij} }$ \change{$\in \SE{3}$} consisting in: the vector $\posW{\nodeij} \in \R{3}$ representing the origin of the node frame  expressed w.r.t. the inertial frame and the node orientation matrix $\rotmW{\nodeij} \in \SO{3}$ composed of unit column vectors expressed w.r.t. the inertial frame. All the node frames  have the same orientation when the mesh is \emph{stretched} (namely, when all the square sections belong to the same plane) -- see, \eg, Fig.\ref{fig:3by3}.
\end{preliminary}

\subsection{System State}

The morphing cover skeleton obtained by applying  the procedure described in Preliminary \ref{preliminary:built-skeleton} is, at all effects and purposes, a mechanical system.  Hence, we have to define the \emph{system state}: namely,
a set of relationships from which the position and  velocity of each point of the cover can be uniquely determined in the inertial frame.
Classically, this problem is solved by choosing a set of minimum variables representing the system configuration and by computing their time derivative  to represent the system velocity. The variables $\statePosRotm, \stateVelRotm \in \R{k}$, for instance, are often used to represent the joint angles and  velocities associated with fixed-base manipulators with $k$ rotational/prismatic joints \cite{roboticsHandbook}. 

Being a \emph{highly} parallel mechanism, the process of finding a set of minimum variables that uniquely characterize the cover configuration may not be a straightforward task. For this reason, the route we follow to characterize the system state is to consider a \emph{complete} variable representation complemented by a set of constraints. More precisely, the system state is represented by the pair $(\statePosRotm,\nodesAbsVel)$, with:

\begin{IEEEeqnarray}{RCL}
    \IEEEyesnumber \phantomsection \label{systemConfiguration}
    \statePosRotm &=& \brRound{ \posW{\node{1}{1}},\rotmW{\node{1}{1}} \cdots \posW{\nodeij},\rotmW{\nodeij} \cdots \posW{\node{n}{m}},\rotmW{\node{n}{m}} } \IEEEeqnarraynumspace 	\IEEEyessubnumber \ \\
    \nodesAbsVel &=& \brRound{ \linVelW{\node{1}{1}}, \angVelW{\node{1}{1}} \cdots \linVelW{\nodeij}, \angVelW{\nodeij} \cdots 
    \linVelW{\node{n}{m}}, \angVelW{\node{n}{m}} }  \IEEEeqnarraynumspace 	\IEEEyessubnumber 
\end{IEEEeqnarray}
complemented by the equations that constrain the nodes to move according to the spherical joints that connect them, \ie
\begin{IEEEeqnarray}{RCL}
    \IEEEyesnumber
    g \brRound{\statePosRotm} &=& \zero{} \IEEEyessubnumber  \\
    \Jc \brRound{\statePosRotm} \nodesAbsVel &=& \zero{}. \IEEEyessubnumber 
\end{IEEEeqnarray}
In other words, the \emph{system state} belongs to the following set:
\begin{equation*}
\resizebox{.98 \columnwidth}{!}{$
    \Theta \coloneqq \brCurly{ \brRound{ \statePosRotm,\nodesAbsVel } \in \SE{3}^{nm} \times \R{6nm}
    :
    g \brRound{\statePosRotm} = \zero{},
    \Jc  \brRound{\statePosRotm} \nodesAbsVel = \zero{} }
$}
\end{equation*}
\change{where $\SE{3}^{nm}$ is the $nm$-ary Cartesian power of the set $\SE{3}$.}
What follows defines the holonomic   and   differential constraints $g \brRound{\statePosRotm} = \zero{}$ and $\Jc  \brRound{\statePosRotm} \nodesAbsVel =  \zero{}$, respectively. The linear map $\Jc  \brRound{\statePosRotm}$ is here referred to as the \emph{constraint jacobian}.

\subsection{Holonomic and Differential Constraints}

To define the holonomic constraints $g \brRound{\statePosRotm} {=} \zero{}$, we need to find out the geometric relationships that spherical joints impose on the motion between two nodes.
Now, considering Fig.~\ref{fig:sphericalJoint2DCreo}, observe that the midpoint $B$ in the spherical joint must satisfy the following constraint: 
\begin{equation*}
    \posW{\nodeij} + \rotmW{\nodeij} \pos{\nodeij}{\nodeij}{B} = \posW{\nodeijp} + \rotmW{\nodeijp} \pos{\nodeijp}{\nodeijp}{B} \: .
\end{equation*}
Note also that  $\pos{\nodeij}{\nodeij}{B}$ and $\pos{\nodeijp}{\nodeijp}{B}$ are constant. The former is equal to $\distJointI = \ev{1} l$, while the latter to $-\distJointI$. Therefore, one has
\begin{equation} \label{eq:Spherical Joint Position}
    \posW{\nodeij} + \rotmW{\nodeij} \distJointI = \posW{\nodeijp} - \rotmW{\nodeijp} \distJointI \: .
\end{equation}
By repeating the above procedure for all spherical joints, and by moving the terms on the right-hand side of \eqref{eq:Spherical Joint Position} to its  left-hand side, we obtain the holonomic constraints $g \brRound{\statePosRotm} {=} \zero{}$.

Now, to obtain the differential constraints $\Jc \brRound{\statePosRotm} \nodesAbsVel = \zero{}$, differentiate 
 \eqref{eq:Spherical Joint Position} \wrt to time, which leads to
\begin{equation} \label{eq:Shperical joint velocity}
\resizebox{.89 \columnwidth}{!}{$
    \linVelW{\nodeij} + \skw{\angVelW{\nodeij}} \rotmW{\nodeij} \distJointI = \linVelW{\nodeijp} - \skw{\angVelW{\nodeijp}} \rotmW{\nodeijp}  \distJointI
    $} .
\end{equation}
Rearranging the terms of \eqref{eq:Shperical joint velocity} in a matrix form leads to
\begin{equation} \label{eq:Shperical joint velocity Matrix form}
\resizebox{.89 \columnwidth}{!}{$
    \bm{ \eye{3} & - \skw{ \rotmW{\nodeij}  \distJointI } }
    \bm{ \linVelW{\nodeij} \\ \angVelW{\nodeij} } =
    \bm{ \eye{3} & \skw{ \rotmW{\nodeijp}  \distJointI } } 
    \bm{ \linVelW{\nodeijp} \\ \angVelW{\nodeijp} }
    $} .
\end{equation}
Using  two selector matrices $\selm{\nodesAbsVel[\nodeij]}$ and $\selm{\nodesAbsVel[\nodeijp]}$, we can express the above constraint w.r.t the velocity $\nodesAbsVel$, namely
\begin{equation} \label{eq:Spherical joint nodesAbsVel}
\resizebox{.89 \columnwidth}{!}{$
    \bm{ \eye{3} & - \skw{ \rotmW{\nodeij}  \distJointI } }
    \selm{\nodesAbsVel[\nodeij]} \nodesAbsVel =
    \bm{ \eye{3} & \skw{ \rotmW{\nodeijp}  \distJointI } } 
    \selm{\nodesAbsVel[\nodeijp]} \nodesAbsVel.
    $}
\end{equation}
An analogous constraint can be found for the spherical joint that connects node $\nodeij$ to $\nodeipj$.
Then, by stacking all constraints in a matrix leads to the differential constraints:
\begin{equation} \label{eq:jacobianRelathionshipFloatingBase}
    \Jc \nodesAbsVel = \zero{}, \ \Jc\in \R{ 3 \numberSphericalJoints \times  6nm }.
\end{equation}

\subsubsection{Fixed Father Node Constraints}
Besides the holonomic and differential constraints due to the cover spherical joints, we add another constraint that simplifies -- without loss of generality -- the overall modeling task: we assume that the father node does not move in the inertial frame. This translates in additional holonomic constraints, \ie, the father node position $\posW{\node{1}{1}}$ and orientation $\rotmW{\node{1}{1}}$ are constant. Thus
\begin{equation} \label{eq:baseFixed}
    \nodesAbsVel[\node{1}{1}] \coloneqq \bm{\linVelW{\node{1}{1}} \\ \angVelW{\node{1}{1}}} \coloneqq \vb = \selm{\nodesAbsVel[\node{1}{1}]} \nodesAbsVel   = 0
\end{equation}
where the selector matrix $\selm{\nodesAbsVel[\node{1}{1}]}$ is used to select only the term $\nodesAbsVel[\node{1}{1}]$ from the vector $\nodesAbsVel$. In the language of the floating base system, \eqref{eq:baseFixed} is the so called \emph{fixed-base} assumption and complements the constraints \eqref{eq:jacobianRelathionshipFloatingBase} leading to
\begin{equation} \label{eq:jacobianRelathionship}
    \Jc \nodesAbsVel = \zero{}, \ \Jc\in \R{ 3 \numberSphericalJoints +6 \times  6nm }.
\end{equation}

\subsubsection{Constraint Jacobian Properties} \label{subse:Jacobian of constraints}
The rows of the constraint jacobian $\Jc$ are the system's differential constraints. Some of these constraints, however, might be redundant, \ie, the jacobian $\Jc$  may be rank deficient. The  jacobian rank indicates the  number of active constraints at a given cover configuration $\statePosRotm$. Once the number of active constraints is identified, we can evaluate the \emph{instantaneous} degrees of freedom ($\dof$) of the morphing cover at a given configuration $\statePosRotm$ as follows:
\begin{equation} \label{eq:DoF computation}
    \dof = \dimV{\nodesAbsVel} - \rank{\Jc}.
\end{equation}
The rank computation of the constraint jacobian $\Jc$ is performed numerically, and its value may change along with the mesh configuration $\statePosRotm$. The constraint jacobian $\Jc$, however, will provide us with important information. For instance, its null space, the so-called $\nullSpaceMatrix[\nodesAbsVel]$%
, represents the vector space of all feasible cover movements in a given configuration.

\subsection{System Modeling in Relative Coordinates}

Constraints \eqref{eq:baseFixed} and \eqref{eq:jacobianRelathionship} are derived with the state vector $\nodesAbsVel$, which represents the node velocities expressed in the \emph{inertial frame}. Remind that nodes are connected via spherical joints. Since our objective is to actuate the spherical joints, then the relative angular velocities between two nodes will be affected by the joint actuation. Therefore, it is important to have a system state representation in terms of \emph{relative node velocities}.

\begin{definition}
    $\nodesRelOmega \in \R{3\numberSphericalJoints}$ is the vector containing all the relative angular velocities between consecutive links, \ie:
    \begin{equation}
    \resizebox{.87 \columnwidth}{!}{$
        \nodesRelOmega = \brRound{
        \angVel{\node{1}{1}}{\node{1}{1}}{\node{1}{2}}
        \, \cdots \, 
        \, \angVel{\nodeij}{\nodeij}{\nodeijp} \, ,
        \, \angVel{\nodeij}{\nodeij}{\nodeipj}
        \, \cdots \, }
        $} .
    \end{equation}
\end{definition}
\begin{definition}
    $\VbAndNodesRelOmega \in \R{6+3\numberSphericalJoints}$ is the vector
    of base velocity $\vb$ and relative velocities $\nodesRelOmega$, \ie,:
    \begin{equation}
        \VbAndNodesRelOmega = \brSquare{\vb  \; \text{;} \;  \nodesRelOmega} \: .
    \end{equation}
\end{definition}

Given the system velocities $\nodesAbsVel$ in the inertial frame, it is straightforward to define a linear transformation $\convm{\VbAndNodesRelOmega}{\nodesAbsVel}$ such that $\VbAndNodesRelOmega = \convm{\VbAndNodesRelOmega}{\nodesAbsVel}\nodesAbsVel$.
The first term $\vb$ is contained in both terms $\nodesAbsVel$ and $\VbAndNodesRelOmega$. All the other terms of $\VbAndNodesRelOmega$ can be computed as
\begin{IEEEeqnarray}{RCL}
    \IEEEyesnumber \phantomsection \label{eq:w=Rw}
    \angVel{\nodeij}{\nodeij}{\nodeijp} &=& \rotm{\nodeij}{\frameW} \angVel{\frameW}{\nodeij}{\nodeijp} \IEEEeqnarraynumspace 	\IEEEyessubnumber \label{eq:w=Rw A} \\
    \angVel{\frameW}{\nodeij}{\nodeijp} &=& \angVelW{\nodeijp} - \angVelW{\nodeij}. \IEEEeqnarraynumspace 	\IEEEyessubnumber \label{eq:w=Rw B}
\end{IEEEeqnarray}
Using the selector matrices $\selm{\angVelW{\nodeij}}$ and $\selm{\angVelW{\nodeijp}}$, one has:
\begin{equation}  \label{eq:Rw}
    \angVel{\frameW}{\nodeij}{\nodeijp} = \selm{\angVelW{\nodeijp}} \nodesAbsVel - \selm{\angVelW{\nodeij}} \nodesAbsVel,
\end{equation}
and substituting \eqref{eq:Rw} into \eqref{eq:w=Rw A} gives:
\begin{equation}
    \angVel{\nodeij}{\nodeij}{\nodeijp} = \rotm{\nodeij}{\frameW} \brRound{ \selm{\angVelW{\nodeijp}} - \selm{\angVelW{\nodeij}}} \nodesAbsVel .
\end{equation}
By repeating this procedure for all the spherical joint, we finally get to a linear mapping between $\VbAndNodesRelOmega$ and $\nodesAbsVel$
\begin{equation} \label{eq:convmV2Nu}
    \VbAndNodesRelOmega = \convm{\VbAndNodesRelOmega}{\nodesAbsVel} \nodesAbsVel .
\end{equation}
We also apply the transformation to the null space projector $\nullSpaceMatrix[\nodesAbsVel]$, so we can pass from the null space projector expressed in the inertial frame to that expressed in relative coordinates:
\begin{equation} \label{eq:convmZv2Znu}
    \nullSpaceMatrix[\VbAndNodesRelOmega] =  \convm{\VbAndNodesRelOmega}{\nodesAbsVel} \nullSpaceMatrix[\nodesAbsVel].
\end{equation}

\begin{figure}[t]
    \centering
    \includegraphics[height=99pt]{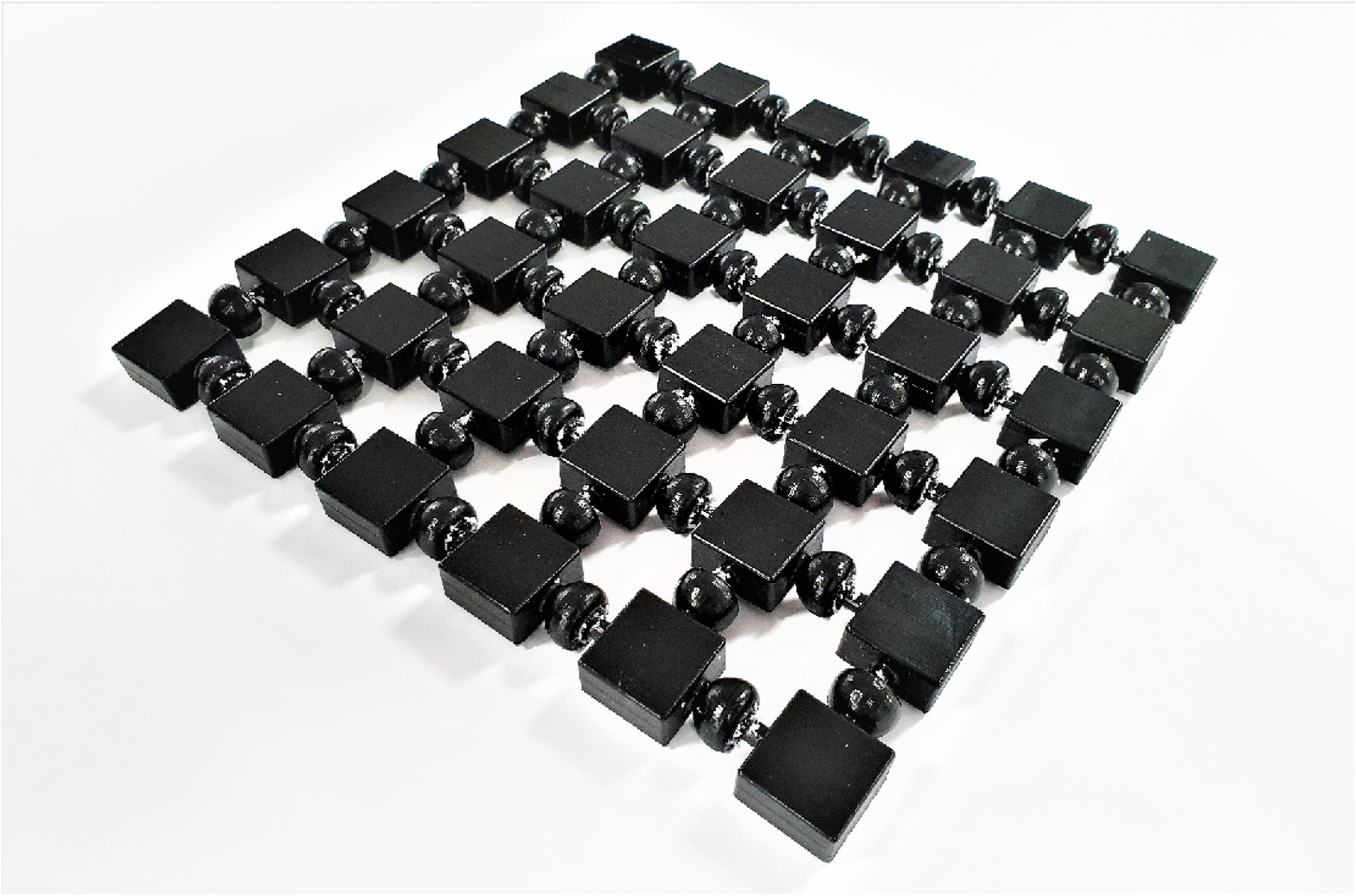}
    \vspace{5.5mm}
    \caption[]{$3$-D printed prototype of a $6\times6$ morphing cover skeleton with passive spherical joints designed with the proposed approach.}
    \label{fig:6by6_real}
\end{figure}

%% file: 4-actuatorSelection.tex
\section{Actuation Positioning} \label{sec:actuatorChoice}

In the previous section, we designed the skeleton of the morphing cover concept, and we addressed the kinematic model that characterizes its evolution.
To actuate the cover, we assume that all or some of the three degrees of freedom ($\dof$) of each spherical joint are  actuated. This section addresses the problem of defining which $\dof$ of the spherical joints shall be actuated.
In general, actuation positioning for parallel structures is not trivial since the number of (actionable) joints is larger than the number of structure $\dof$. Therefore, a possible criterion is to place a number of actuators equal to the number of the cover degrees of freedom, \ie render the mesh \emph{fully actuated}. However, there may exist a large number of actuation patterns that ensure a fully actuated cover, and this number may change depending on the cover motion.  What follows proposes solutions to these motor placement problems using genetic algorithms.

Let us make the following assumptions.

\begin{assumption} \label{ass:motorPosition}
    Motors actuate the $3\text{-}\dof$ spherical joints. Therefore, a maximum of three motors can be placed in correspondence of one spherical joint.
\end{assumption}
In practice, we make Assumption~\ref{ass:motorPosition} true
by assuming that a spherical joint is composed of three intersecting revolute joints. Therefore, we may have less than three motors in correspondence of a spherical joint, and in this case, the non-actuated revolute joints remain \emph{passive}.
Under Assumption~\ref{ass:motorPosition}, the overall objective of the genetic algorithm presented below 
can be stated as follows:

\begin{itemize}
    \item The motors shall actuate a set of revolute joints that ensure  \emph{full actuation} %
    of the cover in a neighborhood of a cover configuration  $\Bar{\statePosRotm}$.
\end{itemize}

An example of the several mechanisms that can actuate the revolute joints is the \emph{Faulhaber} brushless motor 0308H003B ($3$[mm] diameter, $8.4$[mm] length, $0.026$[mNm] holding torque) with a planetary gearheads \emph{Faulhaber} 03A 125:1. We are currently investigating its use and integration in one of the prototypes that we are currently producing, \change{-- see Fig.}~\ref{fig:6by6_real}. 

Since the motors actuate all or some of the three joints composing the spherical joint connecting two nodes,  we make the following actuation assumption.

\begin{assumption} 
    \label{ass:motorAngularDefinition}
    Consider the spherical joint connecting the nodes ${\nodeij}$ and ${\nodeijp}$. Then, a motor placed at this spherical joint imposes one and only one component of the associated relative angular velocity $\angVel{\nodeij}{\nodeij}{\nodeijp}$. 
\end{assumption}
In light of Assumption \ref{ass:motorAngularDefinition}, the problem of positioning the cover motors corresponds to the problem of choosing a set of relative angular velocities that ensures full actuation. 

To address the problem, we make use of the following property of the null space projector $\nullSpaceMatrix[\VbAndNodesRelOmega]$.

\begin{corollary} \label{cor:nullspace}
    The problem of motor positioning 
    is equivalent to the problem of finding 
    the indexes of the linear independent rows of the matrix $\nullSpaceMatrix[\VbAndNodesRelOmega]$ .
\end{corollary}
\noindent
The proof is given in the \hyperlink{labelAppendix}{Appendix}. We now have to find the indexes of the linear independent rows of the matrix~$\nullSpaceMatrix[\VbAndNodesRelOmega]$.

Theoretically, we may address this problem with a \emph{brute force approach}. The number of all possible sets to check is  given by the binomial coefficient:
\begin{equation}
    \label{number-og-possibility-actuation}
    \binom{3 \numberSphericalJoints}{\dof} = \binom{3 \times \text{Number spherical joints}}{\text{Number of motors}}.
\end{equation}
Therefore, the objective is to choose one actuation pattern -- among the numerous possibilities~\eqref{number-og-possibility-actuation} -- that renders the cover fully actuated in a neighborhood of the configuration~$\Bar{\statePosRotm}$.
To give the reader an insight on
the size of the brute force approach, consider a $4 \times 4$ mesh and assuming that we want to place $10$ motors -- \ie a configuration with $10 \dof$. In this case, there are about $10^{11}$ possible actuation patterns to check, among which one has to be chosen. Hence, the brute force approach becomes quickly unusable for large covers, especially when additional logic should be considered when choosing a pattern, \eg, to minimize the number of motors at a spherical joint.

To address this problem, we implement a simple genetic algorithm (GA) based on single-point crossover and with a fitness function $f_{GA}$ to be maximized:
\begin{equation} \label{eq:fitnessGA}
    f_{GA} \brRound{h} =  \bigg| \det{\Big(\nullSpaceMatrix[\VbAndNodesRelOmega]\brRound{h,:}}\Big) \bigg| \, \; f_R \brRound{h} \; f_P \brRound{h},
\end{equation}
where $h$ is a candidate string
that contains the indexes of the selected rows of $\nullSpaceMatrix[\VbAndNodesRelOmega]$. 
The term $f_R \brRound{h}$ is a reward added to privilege solutions that place the lower number of motors in correspondence of a single spherical joint.
The term $f_P \brRound{h}$ is a dynamic penalty cost added to maintain population diversity and create a new generation without duplicate strings.

The output of the GA is a population $\mathcal{P}$ composed by strings $h$. Each string contains the indexes of linearly independent rows of the matrix $\nullSpaceMatrix[\VbAndNodesRelOmega]$, and  -- thanks to~\cref{cor:nullspace}~-- it corresponds to the sets of optimal motor placement. However, the optimality of the solution is only guaranteed at the specific mesh configuration $\Bar{\statePosRotm}$ considered by the GA.

In order to robustify the cover full actuation in a neighborhood of $\Bar{\statePosRotm}$ \change{and mitigate the dependence on the cover geometry}, we perform a sensitivity analysis to check the variation of the determinant \wrt small changes in the mesh configuration. Define $g_{SA}$
\begin{equation} \label{eq:sensitivity analysis}
    g_{SA} \brRound{h} = \sum_ {i \in h} \left( \frac{\partial }{\partial \theta_i} \det{\Big(\nullSpaceMatrix[\VbAndNodesRelOmega]\brRound{h,:}}\Big)  \right)^ 2 ,
\end{equation}
where $\theta_i$ is a small angular variation produced by the $i$-th actuator. We compute $g_{SA}$ for all the strings that belong to the population $\mathcal{P}$. The candidate with lower $g_{SA}$ is the solution whose determinant is less affected by changes in the neighbourhood configuration.
In the ideal case of $g_{SA} = 0$, the determinant remains constant in the neighborhood of the configuration, and the motors remain linear independent.

The above genetic algorithm will be applied to locate the motors of several cover examples in section~\ref{sec:simulations}. More precisely, the outputs of the algorithm that will be used in the sections next are defined by the following statements.

\begin{definition} \label{def:nodesRelOmegaAct}
    The vector $\nodesRelOmegaAct \in \R{\dof}$ represents the relative angular velocities selected using the GA, and therefore, it represents the vector of motor velocities.%
\end{definition}

\begin{definition}
    The submatrix $\nullSpaceMatrix[\text{act}] \in \R{\dof \times \dof}$ is obtained by selecting the linear independent rows of the matrix $\nullSpaceMatrix[\VbAndNodesRelOmega]$.
\end{definition}

In light of the definitions above, we can state the following lemma that is useful to relate the relative angular velocities $\nodesRelOmegaAct$ (motor actions) to the cover velocities $\nu$ and all nodes velocities $ \nodesAbsVel$ expressed in the inertial frame. 

\begin{lemma} \label{lem:fromInputTo}
    Given $\nodesRelOmegaAct$, the absolute and relative cover velocities $\nodesAbsVel$ and $\VbAndNodesRelOmega$ are evaluated as:
    \begin{IEEEeqnarray}{RCL}
        \IEEEyesnumber
        \VbAndNodesRelOmega &=& \nullSpaceMatrix[\VbAndNodesRelOmega] \nullSpaceMatrix[\text{act}]^{-1} \nodesRelOmegaAct , \IEEEeqnarraynumspace 	\IEEEyessubnumber \label{eq:fromInputToRelC} \\
        \nodesAbsVel &=& \nullSpaceMatrix[\nodesAbsVel] \nullSpaceMatrix[\text{act}]^{-1} \nodesRelOmegaAct . \IEEEeqnarraynumspace 	\IEEEyessubnumber \label{eq:fromInputToAbsC}
    \end{IEEEeqnarray}
\end{lemma}

The GA has been developed using the node angular velocities expressed in relative coordinates because of assumption \ref{ass:motorPosition}. It scales, however, to the case where the cover were given with an actuation mechanism that acts directly the nodes state in the inertial frame. This additional feature, however, is not exploited in this article. 

%% file: 5-control.tex
\section{Control Design} \label{sec:control}

In the previous section, we considered a specific cover configuration $\Bar{\statePosRotm}$, and we placed a number of motors equal to the number of the cover degrees of freedom at that configuration. The motor pattern ensures a minimum sensitivity of the cover degrees of freedom, which are, thus, expected not to vary in a (relatively \emph{large}) neighborhood of the cover configuration $\Bar{\statePosRotm}$.
This section presents the control design for the cover motors. 

The overall control objective is to allow the cover to morph (or converge) into a desired shape via the application of proper motor velocities, namely, via the relative angular velocities $\angVel{\nodeij}{\nodeij}{\nodeijp}$ -- see Assumption~\ref{ass:motorAngularDefinition}.
More precisely, the  desired cover shape is given by the mapping $z = f \brRound{x,y,t}$, which induces the normal
$ \versorW{n}{} \brRound{t} ~{=}~\Big. \nabla f \brRound{x,y,t}$. Concerning the cover desired shape,
we also make the following assumption. 
\begin{assumption}
    The desired shape $z = f \brRound{x,y,t}$ passes through the constant origin of the father node $(1,1)$. Namely
    \begin{equation*}
        z_{\node{1}{1}} = f \brRound{ x_{\node{1}{1}}, y_{\node{1}{1}},t } \quad \forall t.
    \end{equation*}
\end{assumption}

Now, consider the node $\nodeij$. The control objective for the node $\nodeij$ is here defined as the  alignment of its normal  $\versorW{z}{\nodeij}~=~\rotmW{\nodeij} \ev{3}$ with the  desired shape normal evaluated as:
\begin{equation*}
    \versorW{n}{\nodeij} \brRound{t} = \Big. \nabla f \brRound{x,y,t} \Big|_{x_{\nodeij},y_{\nodeij}}
\end{equation*}
where $x_{\nodeij}$ and $y_{\nodeij}$ are the instantaneous position coordinates of the 
node $\nodeij$ \wrt the world frame. Note that as soon as the cover moves, the desired normal associated with the node $\nodeij$ varies even if the desired shape does not change versus time.
Concerning the control law to generate an angular velocity $\angVelW{\nodeij}$ that makes $\versorW{z}{\nodeij} = \rotmW{\nodeij} \ev{3}$ converges toward $\versorW{n}{\nodeij} \brRound{t}$, we make use of the following result. 
\begin{proposition}[\cite{PUCCI201572} p.75] 
\label{proposition-convergence-normal}
Assume that $\angVelW{\nodeij} = \angVelW{\nodeij}^\star$ with:
    \begin{equation*}
    \resizebox{.95 \columnwidth}{!}{$    
        \angVelW{\nodeij}^\star = \angVelW{n\nodeij} + \brRound{k_{\nodeij}\brRound{t} + \frac{\dot{\sigma}\brRound{t}}{\sigma\brRound{t}}} \skw{\versorW{z}{\nodeij}} \versorW{n}{\nodeij} + \lambda_{\nodeij} \brRound{t} \versorW{z}{\nodeij}
    $},
\end{equation*}
and $\angVelW{n\nodeij} = \skw{ \versorW{n}{\nodeij} } \dversorW{n}{\nodeij}$ the instantaneous angular velocity of $\versorW{n}{\nodeij}$, $\lambda_{\nodeij} \brRound{\cdot}$ any real valued continuous function, $\sigma\brRound{\cdot}$ any smooth positive real-valued function such that $\inf_t \sigma \brRound{t} > 0$, and $k_{\nodeij}\brRound{\cdot}$ any continuous positive real-valued function such that $\inf_t k_{\nodeij}\brRound{t} > 0$, ensures exponential stability of the equilibrium $\versorW{z}{\nodeij} = \versorW{n}{\nodeij}$ with domain of attraction $\brCurly{ \versorW{z}{\nodeij} : \versorW{z}{\nodeij} {}^{\tr} \versorW{n}{\nodeij} \neq -1 }$  at $t=0$.
\end{proposition}

The tracking of the fictitious angular velocity $\angVelW{\nodeij}^\star$ for all nodes is attempted using an instantaneous optimization problem composed by a quadratic cost function and a set of linear constraints where the motors angular velocity $\nodesRelOmegaAct$ are the optimization variable. The problem writes:
\begin{IEEEeqnarray}{LLL} 
    \IEEEyesnumber \phantomsection \label{eq:QP}
    \text{min}_{\nodesRelOmegaAct} \;
        & \frac{1}{2} \sum_{i=1}^{n} \sum_{j=1}^{m} \Lambda_{\nodeij} \| \angVelW{\nodeij} - \angVelW{\nodeij}^\star \|_2^2 & \IEEEyessubnumber \label{eq:QPcostFunction} \\
    \text{s.t.}:        
	    & \angVelW{\nodeij} = \selm{\angVelW{\nodeij}} \nullSpaceMatrix[\nodesAbsVel] \nullSpaceMatrix[\text{act}]^{-1} \nodesRelOmegaAct & \forall \, i,j  \IEEEyessubnumber \label{eq:QPcontrolVariableRelation} \\
	    & - \nodesRelOmega[max] \le \nodesRelOmegaAct \le \nodesRelOmega[max] & \IEEEyessubnumber \label{eq:QPlimitMaxVelocity} \\
	    & \cos(\alpha) \le \versorW{x}{\nodeij} {}^\tr \versorW{x}{\nodeijp} \le 1 & \forall \, i,j \IEEEyessubnumber \label{eq:QPRoMlimit1} \\
	    & \cos(\alpha) \le \versorW{y}{\nodeij} {}^{\tr} \versorW{y}{\nodeipj} \le 1 & \forall \, i,j \IEEEyessubnumber \label{eq:QPRoMlimit2}
\end{IEEEeqnarray}
The problem \eqref{eq:QP} translates the task of controlling the cover normals thanks to its cost function: it aims at minimizing the difference between the absolute angular velocity of each node and the reference angular velocity $\angVelW{\nodeij}^\star$, which, in turn, ensures convergence in the ideal case of no inequality constraint -- see Proposition~\ref{proposition-convergence-normal}. 
Note that the reference angular velocity $\angVelW{\nodeij}^\star$ depends on the input $\nodesRelOmegaAct$ due to the term $\dversorW{n}{\nodeij}$. This dependence induces an algebraic loop. We solve the algebraic loop by discretizing the term  $\dversorW{n}{\nodeij}$ so that it no longer depends on the angular velocities, but only on the cover configuration. Although there exist other ways for evaluating the desired shape normals that do not induce algebraic loops, the choice above proves to achieve satisfactory overall tracking performances of a desired shape -- see section~\ref{sec:simulations}.  

To \change{regularize} and avoid sharp, possibly discontinuous, variations of the control inputs, 
two extra terms may be added to the cost function and minimized: \change{the norm of $\nodesRelOmegaAct$and} the difference between $\nodesRelOmegaAct$ and its value computed at the previous iteration.

\change{The hard constraint} \eqref{eq:QPcontrolVariableRelation} \change{links the control variable and the absolute angular velocity of each node. While the matrix $\selm{\angVelW{\nodeij}}$ extracts $\angVelW{\nodeij}$ from $\nodesAbsVel$.
The matrix inverse of $\nullSpaceMatrix[\text{act}]$ can be computed using a damping factor to mitigate instabilities caused by singularities.}

The inequalities \eqref{eq:QPlimitMaxVelocity}, \eqref{eq:QPRoMlimit1}, and \eqref{eq:QPRoMlimit2} limit the control input according to hardware limitations, such as the maximum motor speed and the range of motion of the spherical joints.
More precisely, the mechanical collision between the stud and the socket of a spherical joint reduces its range of motion. The movements of the stud, in fact, must belong to a cone with semi-aperture equal to $\alpha$.
Since the axis of the stud is always parallel to one of the unit vectors of the frame axis, limiting the relative movement of that versor is equivalent to limit the movements of the stud.
In case of a spherical joint placed in between links $\nodeij$ and $\nodeijp$, the scalar product between $\versorW{x}{\nodeij}$ and $\versorW{x}{\nodeijp}$ must be higher than $\cos{\alpha}$, as stated in \eqref{eq:QPRoMlimit1}.
Substituting $\versorW{x}{\nodeij} = \rotmW{\nodeij} \ev{1}$ into \eqref{eq:QPRoMlimit1}, we obtain
\begin{equation}
    \cos(\alpha) \le \ev{1} ^\tr \rotm{\nodeij}{\nodeijp} \ev{1} \le 1 \:
\end{equation}
with $\rotm{\nodeij}{\nodeijp}$ estimated using forward Euler integration, \ie,
\begin{equation}
\resizebox{.88 \columnwidth}{!}{$ 
    \rotm{\nodeij}{\nodeijp} %
    = \brSquare{\eye{3} + \skw{\angVel{\nodeij}{\nodeij}{\nodeijp}} dt } \rotm{\nodeij}{\nodeijp}\brRound{k}
$}.
\end{equation}
$\rotm{\nodeij}{\nodeijp}\brRound{k}$ is the measured rotation matrix at iteration $k$, and $dt$ is the time increment.
The same analysis for the spherical joint between the joints $\nodeij$ and $\nodeipj$ leads to~\eqref{eq:QPRoMlimit2}.

\change{The control laws that are obtained as a solution to} \eqref{eq:QP} \change{ensure (quasi) \emph{global} stability as long as $\angVelW{\nodeij} - \angVelW{\nodeij}^\star = \zero{}, \ \forall (i,j)$.
Under this condition, the nodes normals converge toward those of the desired shape \emph{whatever} (but one, \ie, $\versorW{z}{\nodeij} {}^{\tr} \versorW{n}{\nodeij} = -1$) initial condition we set for the morphing cover -- see the stability properties of} \cref{proposition-convergence-normal}\change{. %
The possibility for the control laws to ensure $\angVelW{\nodeij} - \angVelW{\nodeij}^\star = \zero{}, \ \forall (i,j)$ largely depends on the cover mechanical limits and on the actuation abilities to make the skeleton move, either locally or globally.
If we choose the actuators using the algorithm presented in} \cref{sec:actuatorChoice}\change{, then the morphing cover is \emph{locally fully actuated}.
For this reason, the control laws obtained as the solution to} \eqref{eq:QP} \change{ensures local stability under a large variety of conditions and basically independently from the desired (feasible) shape. 

Since the actuation positioning ensures the cover to be locally fully actuated, critical conditions for the control laws occur when the morphing cover is initialized \emph{far} from the desired shape. %
In these cases, when converging toward the desired shape, the cover may lose \emph{full actuation} leading to ill-posed, singular control laws.
To mitigate these singularity issues and obtain control laws that are globally defined, 
regularization terms can be added in cost functions} \eqref{eq:QPcostFunction}, \change{which can be implemented also by damping the inverse}  $\nullSpaceMatrix[\text{act}]$ in \eqref{eq:QPcontrolVariableRelation}.
\change{Although these regularization terms alter the stability properties of the  controller, they lead to globally defined laws at the cost of higher tracking errors.}

\change{Another strategy that may be attempted when the cover is initialized \emph{far} from the desired shape $\mathcal{A}$ is that of generating a time-varying desired shape $\Bar{\mathcal{A}}(t)$ that starts close to $\statePosRotm_0$ and converges toward $\mathcal{A}$. The desired time-varying trajectory $\Bar{\mathcal{A}}(t)$ shall ensure that the cover degrees of freedom do not change along itself. This process of generating well-posed trajectories may be addressed by an additional control layer, namely a \emph{trajectory planning} layer, that feeds the control laws presented above. The trajectory planning problem, however, is beyond the scope of this article all the more so because it calls for the development of \emph{ad hoc} numerical techniques that can be seldom used on-line}~\cite{dafarra2020}.

%% file: 6-simulation.tex
\section{Simulation Results} \label{sec:simulations}
In this section, we first briefly discuss the implementation of the simulation environment and the results obtained using the motor positioning algorithm presented in \cref{sec:actuatorChoice}. Then, we show simulation results of \change{four} different meshes controlled with the control framework presented in \cref{sec:control}, \change{and finally, we discuss the obtained results}.
The investigated meshes are of dimensions $3 \times 3$, \change{$4 \times 8$}, $8 \times 8$, and $20 \times 20$.
\change{The mesh $4 \times 8$ reproduces the \emph{iRonCub} thigh cover integrating the morphing capability.}
\change{The video of the simulations described in this section are shown in the accompanying movie \url{https://youtu.be/kMfXb2xqGn4?t=350}.}
The code to reproduce the results is available at \urlPaper{}.

\subsection{Simulation Environment}
The approach is tested using a custom-made MATLAB environment that integrates the cover kinematics and its constraints \eqref{eq:jacobianRelathionship}. The environment makes use of a vector representation of the system configuration space \eqref{systemConfiguration}. More precisely, we define the following.

\begin{definition}
    The state space $\statePosQuat$ and its time derivative  is:
    \begin{IEEEeqnarray}{RCL}
        \IEEEyesnumber
        \statePosQuat &=& \brRound{
            \posW{\node{1}{1}} \text{,} \quatW{\node{1}{1}} \cdots
            \posW{\nodeij} \text{,} \quatW{\nodeij} \cdots
            \posW{\node{n}{m}} \text{,} \quatW{\node{n}{m}}}, \IEEEeqnarraynumspace 	\IEEEyessubnumber \\
        \stateVelQuat &=& \brRound{
            \linVelW{\node{1}{1}} \text{,} \dquatW{\node{1}{1}} \cdots
            \linVelW{\nodeij} \text{,} \dquatW{\nodeij} \cdots
            \linVelW{\node{n}{m}} \text{,} \dquatW{\node{n}{m}}}. \IEEEeqnarraynumspace 	\IEEEyessubnumber
    \end{IEEEeqnarray}
\end{definition}

The orientation of the nodes is parameterized using quaternions to avoid singularities generated employing Euler angles.

The system evolution is obtained integrating $\stateVelQuat$ using MATLAB variable step numerical integrator \emph{ode45}. 
The optimization problem \eqref{eq:QP} is solved every $0.01$[s] using OSQP~\cite{osqp} via CasADi~\cite{casadi}. Once motors velocity $\nodesRelOmegaAct$ are known, $\stateVelQuat$ is computed using \eqref{eq:fromInputToAbsC} and the quaternion derivative definition, plus an additional term to compensate numerical integration errors and ensure $\| \quat{}{} \|_2 = 1$ (see \cite{Baumgarte} for details).
The code to model and simulate the system evolution of a multibody system with rigid links relying on a maximal coordinate approach is available at \urlMystica{}.

The mesh geometries and characteristics are listed in \cref{tab:Mesh Characteristics}.

\begin{table}[b]
\caption{Mesh dimensions and hardware limits
}
	\centering
	\begin{tabular}{l c|c}
		\toprule
		\multicolumn{2}{c}{Characteristics}     & Value \\
		\midrule
		\rowcolor{gray!15}
		Node: square length & $L$  & $25$[mm] \\
		Node: distance from square center to $B$ & $l$ & $25$[mm] \\
		\rowcolor{gray!15}
		Motor: Limit angular Velocity & $\nodesRelOmega[max]$ & $5$ $\brSquare{\sfrac{\text{deg}}{\text{s}}}$   \\
		Joints: Limit range of motion & $\alpha$ & $50$[deg]  \\
		\bottomrule
	\end{tabular}
	\label{tab:Mesh Characteristics}
\end{table}

\subsection{Actuation Positioning}
\begin{figure}
        \begin{subfigure}[b]{0.49\columnwidth}
            \centering
            \includegraphics[width=\textwidth]{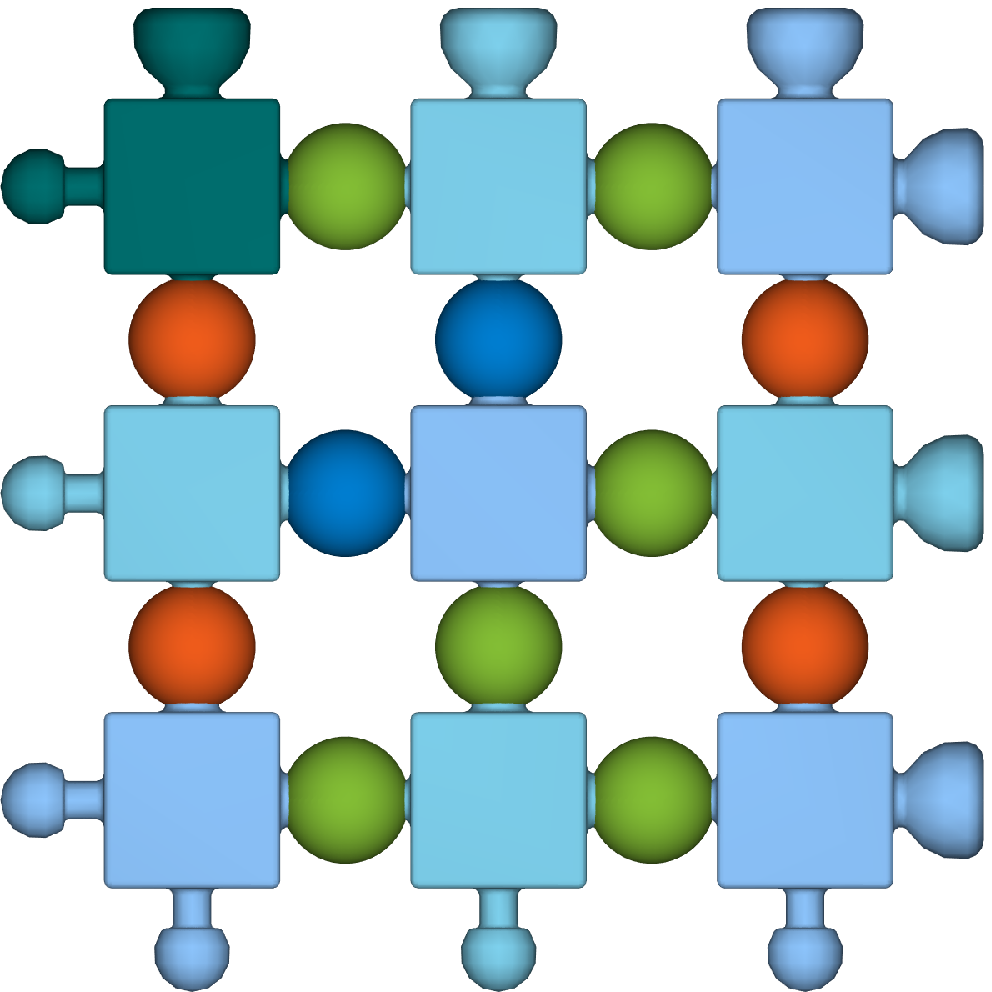}
            \caption[]{$3 \times 3$ mesh}
            \label{fig:sim1_GA}
        \end{subfigure}
        \hfill
        \begin{subfigure}[b]{0.49\columnwidth}  
            \centering 
            \includegraphics[width=\textwidth]{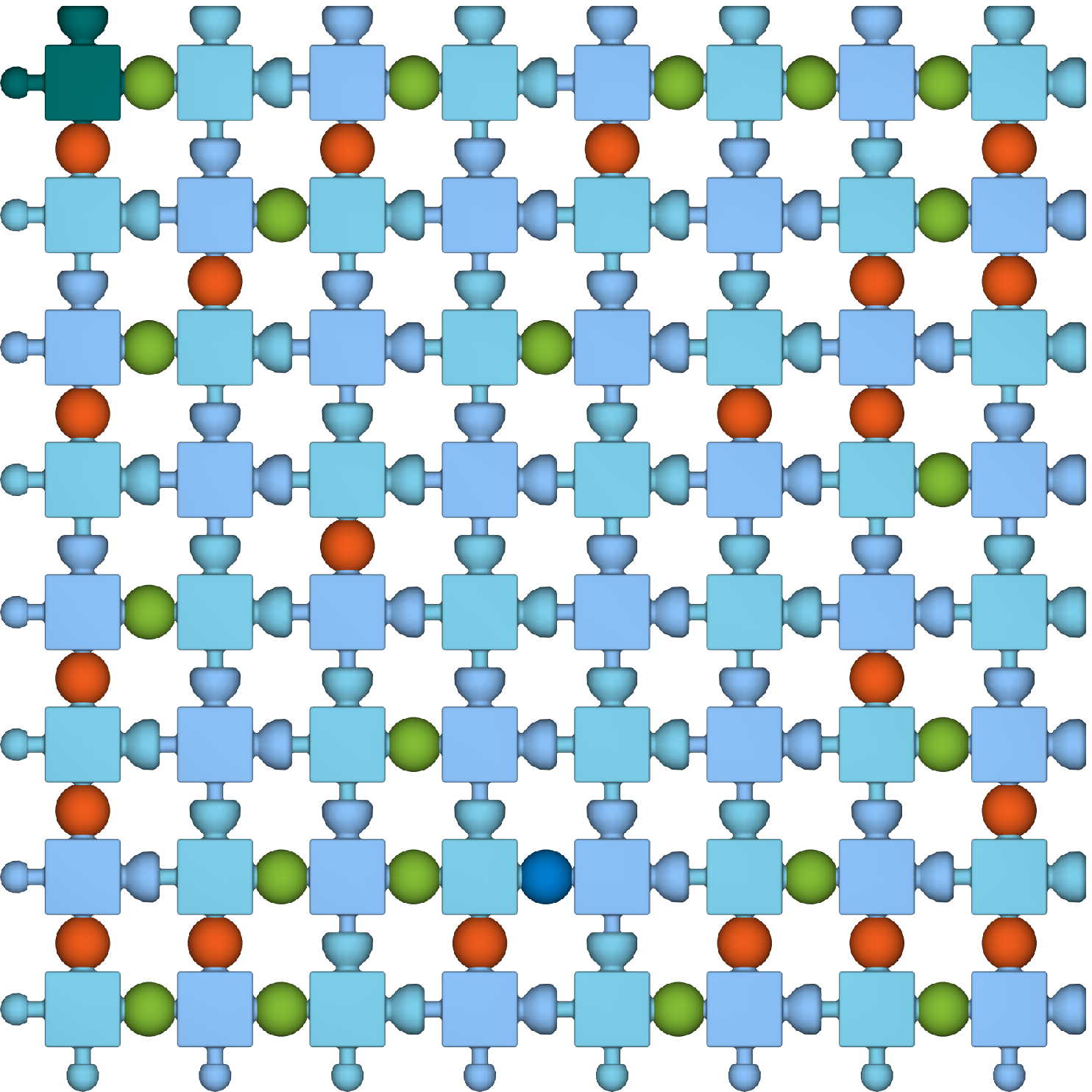}
            \caption[]{$8 \times 8$ mesh}
            \label{fig:sim2_GA}
        \end{subfigure}

        \caption[]{Output of the genetic algorithm described in \cref{sec:actuatorChoice}. The sphere color identifies the presence of a motor. When the color is \emph{sky blue}, the spherical joint is passive.
        If the color is either \emph{red}, \emph{green}, or \emph{blue},  the corresponding axis of rotation is actuated following the RGB convention, while the others axes remain passive.}
        \label{fig:GA}
\end{figure}

The algorithm presented in \cref{sec:actuatorChoice} optimizes the motor placement at a given cover configuration $\Bar{\statePosRotm}$. We ran the algorithm using the cover initial configuration. \change{The cover initial configuration is obtained sampling a \emph{paraboloid} curve for the $3 \times 3$, $8 \times 8$, and $20 \times 20$ meshes, and a \emph{cylinder} for the $4 \times 8$ mesh -- see }\cref{fig:iRonCub-rest}.
Once the configuration $\Bar{\statePosRotm}$ is known, thanks to \eqref{eq:DoF computation}, it is possible to compute the number of $\dof$s and, consequently, the number of motors. At the chosen initial configuration, the meshes $3 \times 3$, \change{$4 \times 8$}, $8 \times 8$, and $20 \times 20$ have $12$, \change{$21$}, $42$, and $114$ $\dof$s respectively.
To initialize the genetic algorithm, we created a random population of $100$ candidate strings, while the crossover and mutation probability are set equal to $0.6$ and $0.01$ after a tuning process started from the values adopted in literature \cite{choosingParametersGA}.
\change{The evolutionary algorithm stops once the population is not evolving over the last $10^3$ generations and the fitness function is higher than a desired threshold.}
\change{The algorithm converged in $10^4$ generations for the meshes $3 \times 3$, $4 \times 8$, and $8 \times 8$ and $10^6$ generations for the mesh $20 \times 20$.}
To compute \eqref{eq:sensitivity analysis}, we considered all the configurations generated applying a motor variation of $\pm 5$[deg] from the initial state. In this way, the cover configuration stays in a neighborhood of the initial configurations. 
Fig. \ref{fig:GA} shows the optimal motor pattern after the sensitivity minimization for $3 \times 3$ and $8 \times 8$. %
Notice the effectiveness of the term $f_R \brRound{h}$ in \eqref{eq:fitnessGA}: the proposed solutions actuate at most one axis of rotation per spherical joint.

\subsection{Control}

Simulations with \change{four} meshes of different sizes are performed to test the controller and validate the ability of the considered meshes to morph into the desired shapes. To test the robustness of the proposed control approach, we also present simulations with noise in the system state used by the control action. 
To measure the control performances, instead, we defined two metrics.
The first  is the angular error in the alignment between the node normal $\versorW{z}{\nodeij}$ and the desired normal $\versorW{n}{\nodeij}$. 
For the node $\nodeij$, the error is defined as:
\begin{equation*}
    \normalAlignmentError{\nodeij} = \left| \arccos{\brRound{ \versorW{z}{\nodeij} {}^{\tr} \versorW{n}{\nodeij}  }} \right|.
\end{equation*}
The minimization of $\normalAlignmentError{\nodeij}$ is obtained thanks to the task implemented in the cost function \eqref{eq:QPcostFunction}. 

The second metric is the node position error, which corresponds to the $z$-distance between the nodes and the curve, \ie,
\begin{equation*}
    \nodePositionError{\nodeij} = \left| \ev{3}^{\tr} \posW{\nodeij} - f \brRound{x_{\nodeij},y_{\nodeij}} \right|.
\end{equation*}
Although the position error is not minimized by the optimization problem \eqref{eq:QP},  it is an indicator to understand how far the cover is from the desired shape.
In the next subsections, we present some data and plots analyzing the two metrics, while meshes simulations are shown in the accompanying video.

\change{The matrix inverse of the constraint} \eqref{eq:QPcontrolVariableRelation} \change{is approximated with a damped pseudoinverse to mitigate possible effects due to singularities. In detail, $\nullSpaceMatrix[\text{act}]^{-1} \simeq \brRound{\nullSpaceMatrix[\text{act}]^{\tr} \nullSpaceMatrix[\text{act}] + 10^{-6} \eye{}}^{-1} \nullSpaceMatrix[\text{act}]^{\tr}$.}

\subsubsection{\textbf{Mesh 3x3}} \label{subsec:3x3}
\begin{figure*}
        \begin{subfigure}[b]{0.32\linewidth}  
            \centering
            \includegraphics[width=\textwidth]{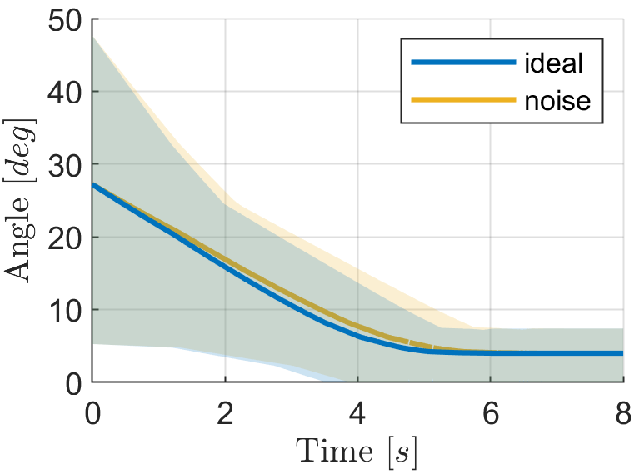}
            \caption{Error alignment normal vectors $\normalAlignmentError{\nodeij}$}
            \label{fig:sim1_AOE}
        \end{subfigure}
        \hfill
        \begin{subfigure}[b]{0.32\linewidth}  
            \centering 
            \includegraphics[width=\textwidth]{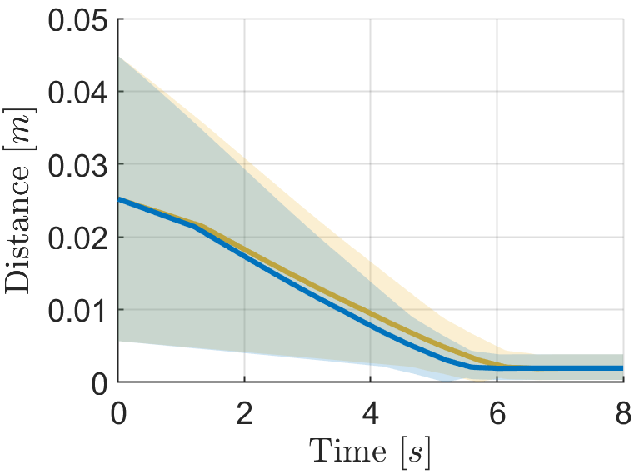}
            \caption{Node position error $\nodePositionError{\nodeij}$}    
            \label{fig:sim1_APE}
        \end{subfigure}
        \hfill
        \begin{subfigure}[b]{0.32\linewidth}  
            \centering 
            \includegraphics[width=\textwidth]{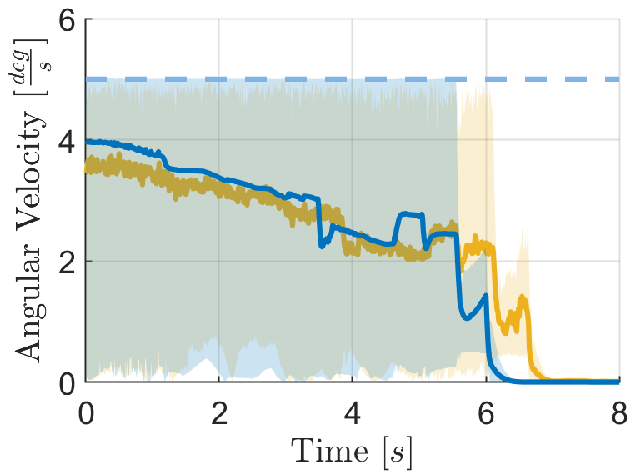}
            \caption{Amplitude Motor Speed $|\nodesRelOmegaAct|$}    
            \label{fig:sim1_OmegaAct}
        \end{subfigure}
        \caption[]{Performance %
        comparison of an ideal simulation and one with artificial noise in the actuation. The test is performed with $3\text{x}3$ mesh. 
        The solid lines represent the average values, while the transparent regions are the $10\text{-}90$th percentile range. In (c), the dashed line is the maximum motor speed $\nodesRelOmega[max]$.
        } 
        \label{fig:sim1}
\end{figure*}

The desired shape that the cover has to track is a function properly scaled and shifted of the form:
\begin{equation*}
   f \brRound{x,y} = x y \cos{y}.
\end{equation*}
In \cref{fig:sim1}, we compare two cases: the first, called \emph{ideal}, where the actuation is considered perfect; the second, called \emph{noise}, where we intentionally added Gaussian noise to \textbf{motor velocity}. The noise is generated such that it can only decrease motor speed to emulate friction \change{or the disturbances generated by the elastic soft membrane}. Its amplitude is proportional to motor speed and it can reach maximum $20\%$ of motor velocity.

After \change{$8$}[s], both the \emph{ideal} and \emph{noise} cases decrease the average orientation error and average position error to $4$[deg] and \change{$2$[mm]}. 
The presence of actuation noise increases the response time without interfering with the task accomplishment. The high-frequency variations in the motor velocity depicted in \cref{fig:sim1_OmegaAct} are due to the Gaussian noise, however, the output of the optimization problem is less oscillating.

\subsubsection{\textbf{Mesh 8x8}} \label{subsec:8x8}
It is tested with a time-varying desired shape properly scaled and shifted in the form of
\begin{equation*}
    f \brRound{x,y,t} =  \cos{\brRound{x + t}} + \cos{\brRound{x + y + t}}.
\end{equation*}
As done for the previous mesh, we compare two simulations: the first one, called \emph{ideal}, characterized by a state that is perfectly estimated, while the second one, called \emph{noise}, because we intentionally added Gaussian noise in the estimated \textbf{state} used by the controller. 
The noise is generated guaranteeing the satisfaction of \eqref{eq:jacobianRelathionship}; in particular, it is applied directly in the actuated joints. However, owing to the closed kinematics, the variations spread in all the joints with a remarkable change in the configuration.
The noise effect is added to emulate measurement errors in the encoders. The maximum amplitude applied to actuated joints was set equal to $0.05$[deg].

\begin{figure*}
        \begin{subfigure}[b]{0.32\linewidth}  
            \centering
            \includegraphics[width=\textwidth]{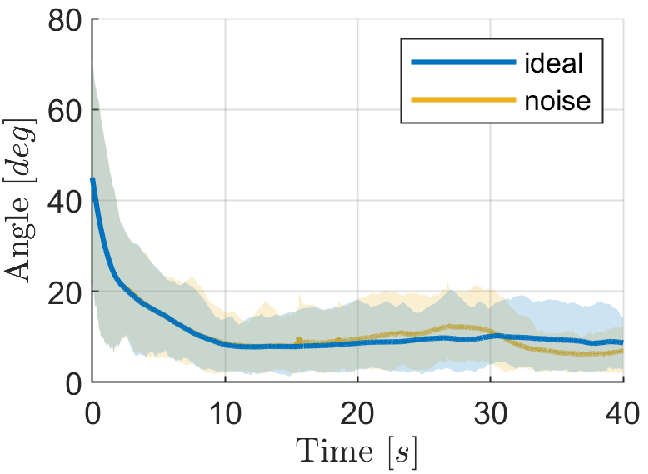}
            \caption{Error alignment normal vectors $\normalAlignmentError{\nodeij}$}
            \label{fig:sim2_AOE}
        \end{subfigure}
        \hfill
        \begin{subfigure}[b]{0.32\linewidth}  
            \centering 
            \includegraphics[width=\textwidth]{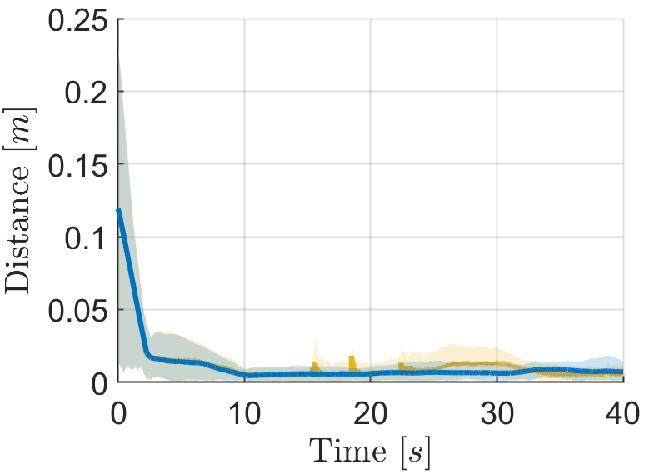}
            \caption{Node position error $\nodePositionError{\nodeij}$}    
            \label{fig:sim2_APE}
        \end{subfigure}
        \hfill
        \begin{subfigure}[b]{0.32\linewidth}  
            \centering 
            \includegraphics[width=\textwidth]{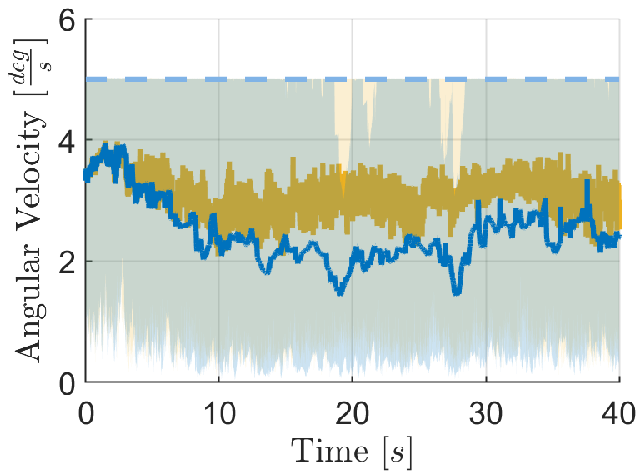}
            \caption{Amplitude Motor Speed $|\nodesRelOmegaAct|$}    
            \label{fig:sim2_OmegaAct}
        \end{subfigure}
        \caption[]{Performance %
        comparison of an ideal simulation and one with artificial noise in the state. The test is performed with $8 \times 8$ mesh.
        The solid lines represent the average values, while the transparent regions are the $10\text{-}90$th percentile range. In (c), the dashed line is the maximum motor speed $\nodesRelOmega[max]$.
        } 
        \label{fig:sim2}
\end{figure*}

Fig.\ref{fig:sim2} shows $40$[s] of simulation; both the \emph{ideal} and \emph{noise} cases decrease the average orientation error and average position error to \change{$9$}[deg] and \change{$8$}[mm].
The presence of state noise increases response time and causes oscillations (see \cref{fig:sim2_OmegaAct}). Yet, the mesh achieves the morphing task with \emph{acceptable} tracking performances according to the aforementioned metrics.

\subsubsection{\textbf{Mesh 20x20}} \label{subsec:20x20}
\begin{figure*}
        \begin{subfigure}[b]{0.32\linewidth}  
            \centering
            \includegraphics[width=\textwidth]{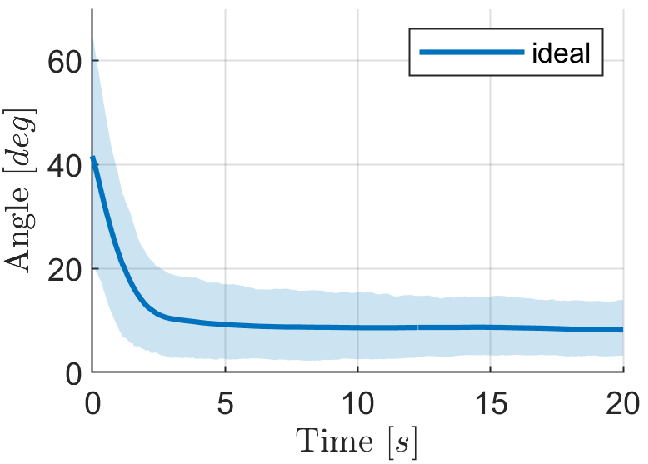}
            \caption{Error alignment normal vectors $\normalAlignmentError{\nodeij}$}
            \label{fig:sim3_AOE}
        \end{subfigure}
        \hfill
        \begin{subfigure}[b]{0.32\linewidth}  
            \centering 
            \includegraphics[width=\textwidth]{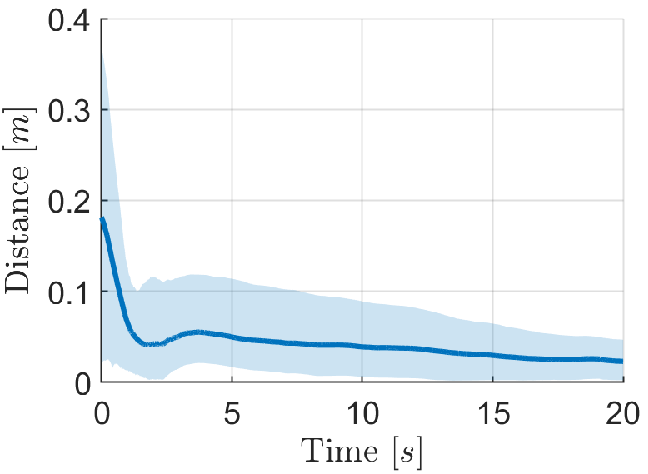}
            \caption{Node position error $\nodePositionError{\nodeij}$}    
            \label{fig:sim3_APE}
        \end{subfigure}
        \hfill
        \begin{subfigure}[b]{0.32\linewidth}  
            \centering 
            \includegraphics[width=\textwidth]{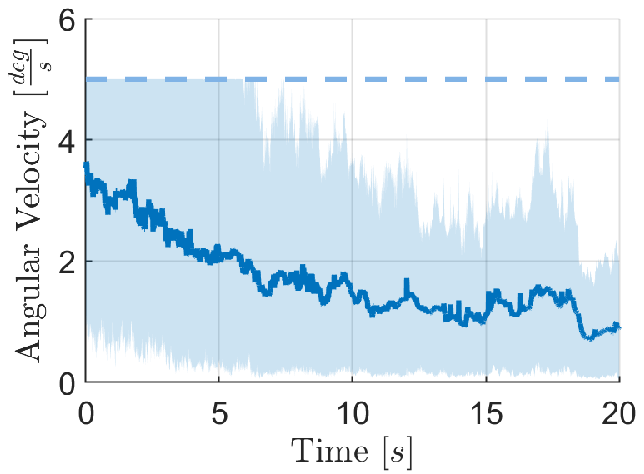}
            \caption{Amplitude Motor Speed $|\nodesRelOmegaAct|$}    
            \label{fig:sim3_OmegaAct}
        \end{subfigure}
        \caption[]{Performance %
        of an ideal simulation executed with mesh $20 \times 20$.
        The solid lines represent the average values, while the transparent regions are the $10\text{-}90$th percentile range. In (c), the dashed line is the maximum motor speed $\nodesRelOmega[max]$.
        }     
        \label{fig:sim3}
\end{figure*}

It is tested with a steady desired shape properly scaled and shifted in the form of
\begin{equation*}
    f \brRound{x,y} = x^2 - y^2.
\end{equation*}
This simulation permitted us to check the capability of the controller to deal with meshes composed of a higher number of nodes; in particular, the number of links is $400$, the number of spherical joints is $760$, and the dimension of the state space $\statePosQuat$ is $2800$.
The problem complexity increases remarkably the computational time; thus, we decided to analyze only the case with ideal actuation and ideal estimation.
Fig.\ref{fig:sim3} shows \change{$20$[s]} of simulation where the mesh passes from an average initial error of \change{$41$[deg]} and \change{$180$[mm]} to \change{$8$[deg]} and \change{$23$[mm]}. Even here, the mesh achieves the morphing task with \emph{acceptable} tracking performances according to the aforementioned metrics; however, \change{$20$[s]} of simulation took \change{$2$[h]} using a PC with Intel Xeon Gold 6128 $3.40$GHz and RAM $128$GB.

\subsubsection{\textbf{Mesh 4x8}} \label{subsec:4x8}
\begin{figure*}
        \begin{subfigure}[b]{0.32\linewidth}  
            \centering
            \includegraphics[width=\textwidth]{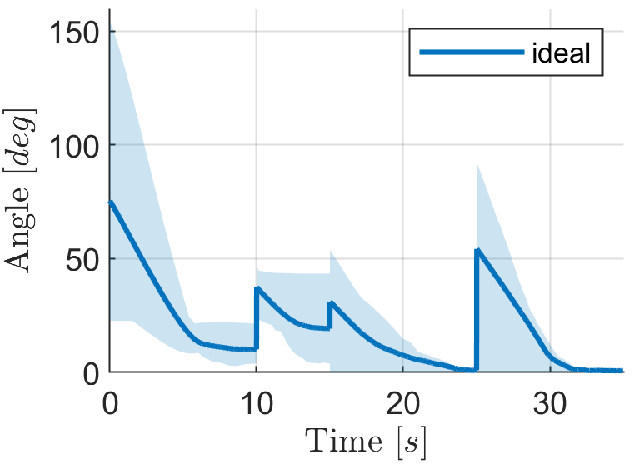}
            \caption{Error alignment normal vectors $\normalAlignmentError{\nodeij}$}
            \label{fig:sim4_AOE}
        \end{subfigure}
        \hfill
        \begin{subfigure}[b]{0.32\linewidth}  
            \centering 
            \includegraphics[width=\textwidth]{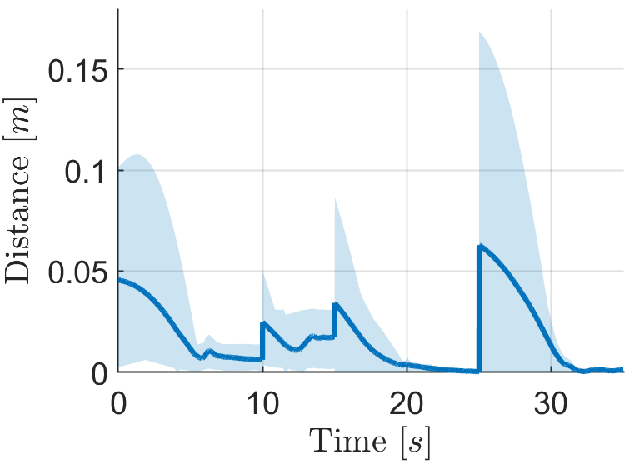}
            \caption{Node position error $\nodePositionError{\nodeij}$}    
            \label{fig:sim4_APE}
        \end{subfigure}
        \hfill
        \begin{subfigure}[b]{0.32\linewidth}  
            \centering 
            \includegraphics[width=\textwidth]{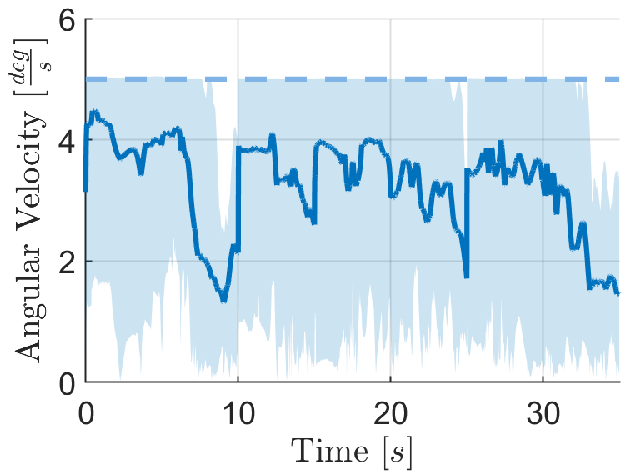}
            \caption{Amplitude Motor Speed $|\nodesRelOmegaAct|$}    
            \label{fig:sim4_OmegaAct}
        \end{subfigure}
        \caption[]{\change{Performance
        of an ideal simulation executed with mesh $4 \times 8$.
        The solid lines represent the average values, while the transparent regions are the $10\text{-}90$th percentile range. In (c), the dashed line is the maximum motor speed $\nodesRelOmega[max]$.}
        }     
        \label{fig:sim4}
\end{figure*}

\begin{figure}
        \begin{subfigure}[b]{0.49\columnwidth}
            \centering
            \includegraphics[trim=50 50 50 50, clip, width=\textwidth]{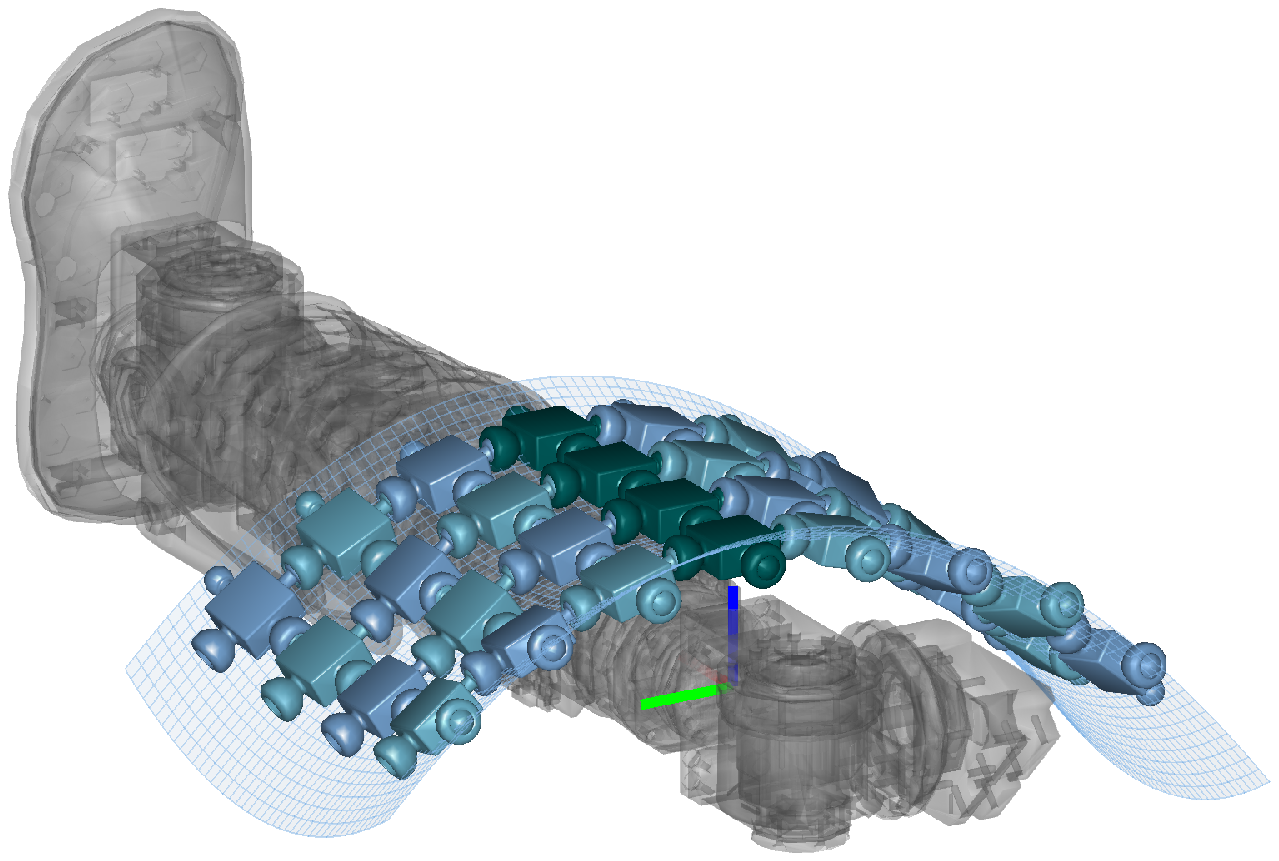}
            \caption[]{$t=10$[s]}
            \label{fig:iCubThigh_t10}
        \end{subfigure}
        \hfill
        \begin{subfigure}[b]{0.49\columnwidth}  
            \centering 
            \includegraphics[trim=50 50 50 50, clip, width=\textwidth]{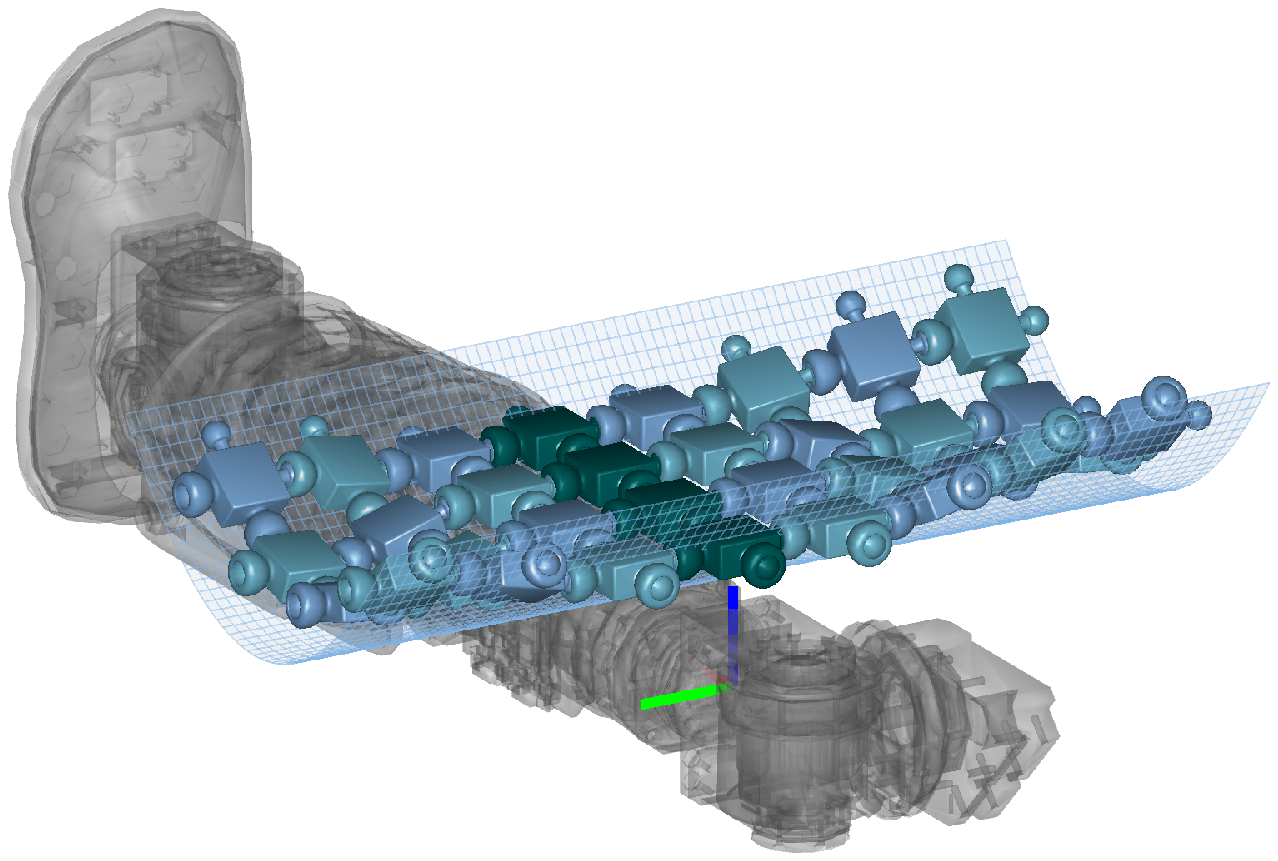}
            \caption[]{$t=15$[s]}
            \label{fig:iCubThigh_t15}
        \end{subfigure}
        \vskip\baselineskip
        \begin{subfigure}[b]{0.49\columnwidth}  
            \centering 
            \includegraphics[trim=50 50 50 50, clip, width=\textwidth]{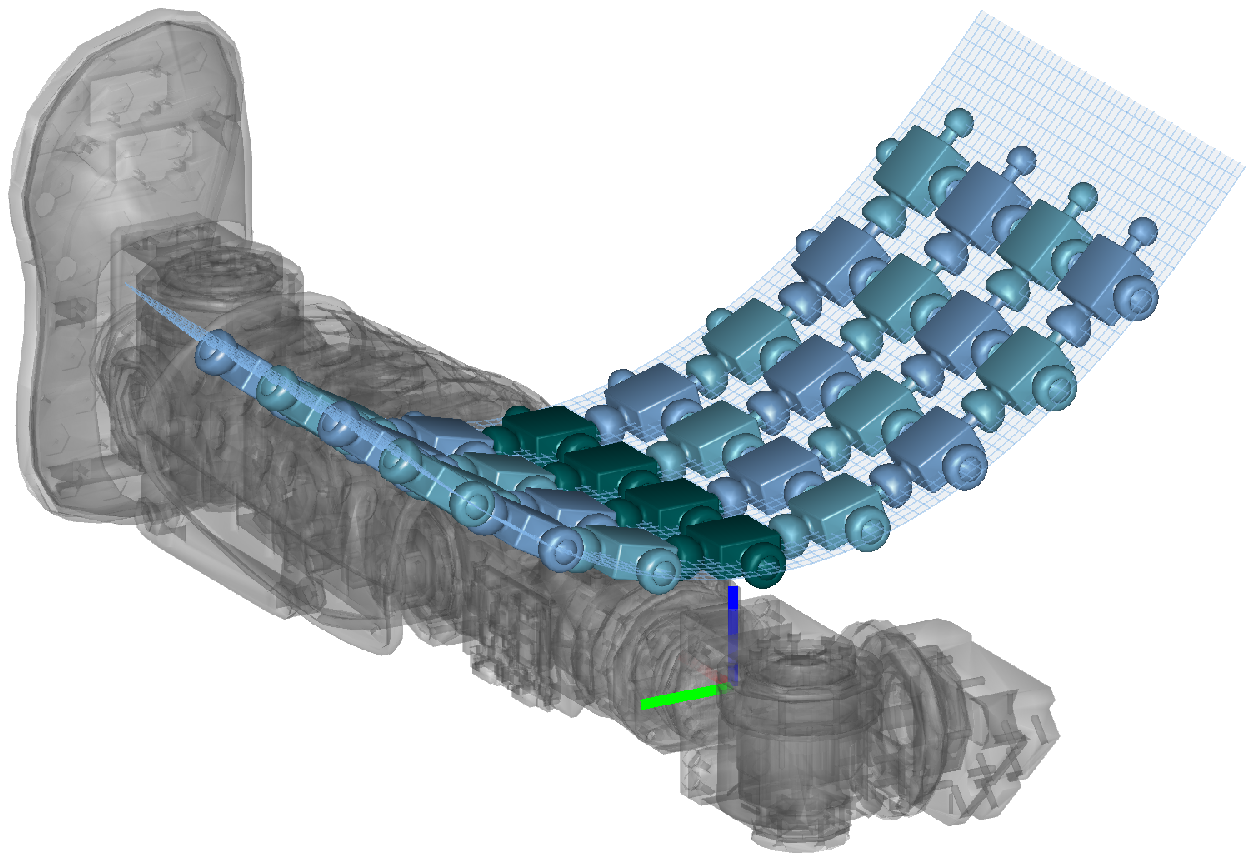}
            \caption[]{$t=25$[s]}
            \label{fig:iCubThigh_t25}
        \end{subfigure}
        \hfill
        \begin{subfigure}[b]{0.49\columnwidth}  
            \centering 
            \includegraphics[trim=50 50 50 50, clip, width=\textwidth]{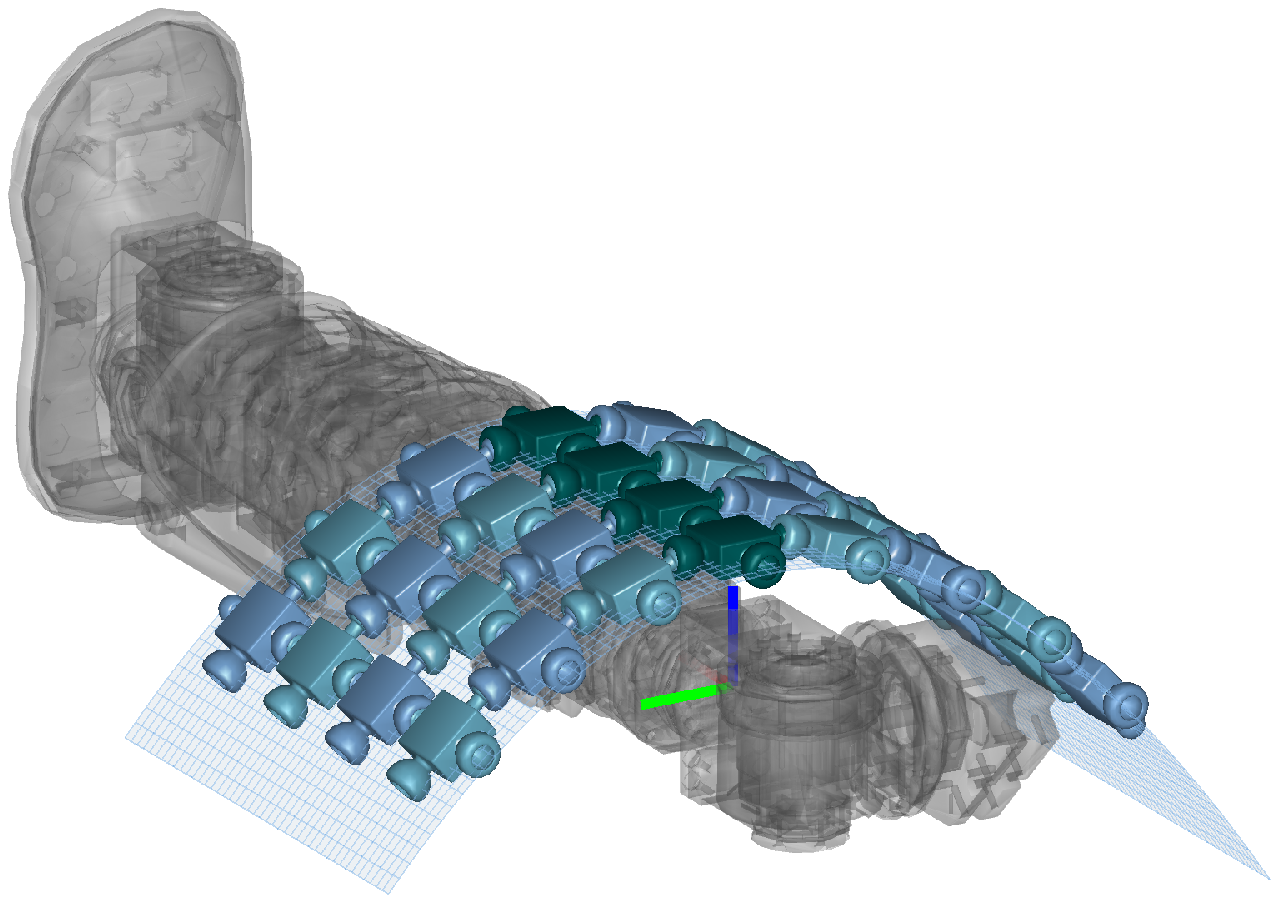}
            \caption[]{$t=35$[s]}
            \label{fig:iCubThigh_t35}
        \end{subfigure}        

    \caption[]{\change{Instants of the $4\times8$ cover movement discussed in }\cref{subsec:4x8}. \change{The four green nodes are rigidly fixed to the leg. The light blue surface is the desired shape. The simulation video is attached in the accompanying material} \url{https://youtu.be/kMfXb2xqGn4?t=482}.}
    \label{fig:iCubThigh}

\end{figure}

\change{
This test presents the conceptual idea of integrating a morphing cover in a humanoid robot. We replaced the \emph{iRonCub} thigh cover with a $4\times8$ mesh. Differently from previous experiment, we rigidly constrained four nodes. The initial configuration is a cylinder (see } \cref{fig:iRonCub-rest}\change{). The mesh is tested with a piecewise function properly scaled and shifted in the form of}
\begin{equation*}
    f \brRound{x,y,t} =
    \begin{cases}
        x^2-y^2 & t \leq 10, \\
        x^2 & 10 < t \leq 15, \\
        y^2 & 15 < t \leq 25, \\
        -y^2 & t > 25.
    \end{cases}
\end{equation*}
\change{Fig.}\ref{fig:sim4}\change{ shows $35$[s] of an ideal simulation without noise. 
The controller reacts to piecewise function discontinuities emulating four different shapes (see }\cref{fig:iCubThigh})\change{. 
The \emph{paraboloids} $\pm{y^2}$ are reproduced with an average orientation error lower than $1$[deg] and average position error of $1$[mm].
Instead, the tracking of the function $x^2-y^2$ is achieved with $10$[deg] and $6$[mm] error. 
Differently, the mesh was not able to copy the \emph{paraboloids} $x^2$, \ie $20$[deg] and $17$[mm] of error, mainly because the \emph{four} fixed nodes constrained a direction of motion.}

\subsection{Discussion}

\change{Above, we have shown the performances of the proposed approach 
in four scenarios, each characterized by a different cover skeleton size, desired shape, and noise source.
The obtained results are satisfactory since the cover tracks a desired time-varying surface with a node-normal alignment error %
less than $10$ [deg] for most of the analyzed shapes.
The presented simulations tend to show that the 
node-normal alignment task allows us to indirectly minimize the node position errors %
to a value lower than the node square length $l$ ($25$ [mm]) in most cases.  Let us remark, however, that the control laws obtained as solutions to} \eqref{eq:QP} \change{do not, in general, imply a minimization of the node position errors.}

\change{The control problem formulated in }\cref{sec:control} \change{moves the skeleton to emulate a continuous function without specifying the desired node coordinates or where the mesh should lie.
On the one hand, this simplifies the control process since it is not required to tessellate the desired shape and estimate the node coordinates. On the other hand, the tracking of a complex desired shape with multiple curvatures could be hard to achieve since the mesh might emulate the desired shape in a coordinate range different from the expected one.
With the current control strategy, we can only select the fixed-node coordinates, but the skeleton might rotate around the fixed-node for large meshes. This behavior slightly appeared with a  $20\times20$ mesh (see the accompanying movie for the simulation video), where the skeleton started to ``slide'' on the desired shape pushed by the minimization of the node-normal alignment.
A desired surface parameterization could be beneficial discouraging mesh movement along unwanted directions.}

\change{The desired shape selection is constrained by physical limitation of the mesh, like the range of motion limits or the position of fixed nodes. The desired shape must belong to a neighbourhood of the mesh configuration, otherwise the controller cannot ensure convergence, unless a trajectory planning layer is introduced.
The mathematical definition of the desired shape is also another drawback since the control problem is only able to track mathematical functions, \ie, binary relations between \emph{domain} and \emph{codomain} that relate each element of the \emph{domain} to exactly one element of the \emph{codomain}. Consequently, owing to a representation reason, the mesh is not able to morph into \emph{non-surjective} functions.}

\change{Concerning the motor positioning algorithm, its result is an optimal solution in a configuration close to the initial one. However the full-actuation is not guaranteed in the whole working space. This geometric dependence does not affect the associated performances since we can assume to work in a configuration near the starting one in most cases. In fact, in all the presented simulations, the meshes did not experience under-actuation and singularity issues.}

%% file: 7-conclusions.tex
\section{Conclusions} \label{sec:conclusions}

This article proposed a  solution for providing  humanoid robots with adaptive morphology capabilities. We presented a novel type of humanoid robot covers whose shape can be actively controlled. 
The proposed morphing cover concept is composed of a skeleton and an elastic membrane. We showed how the skeleton can be obtained as a repetition of basic elements connected by spherical joints, which thus create a parallel mechanism. The design is particularly robust from the mechanical perspective. 
To validate its performance, we carried out extensive simulations with the objective for the cover to morph into desired shapes. 
\change{The mesh evolution is governed by a kinematic controller that orients the node-normals and indirectly minimizes the node-position errors.
This article also proposed a motor placement strategy aiming to minimize the number of motors without losing \emph{full-actuation}. The optimality is assured in a local configuration, however during the presented simulations the meshes did not experience under-actuation issues.}

The choice of spherical joints allows a wide range of movements. However, it increases the complexity of the mechanical design.
\change{Other limitations are the unmodeled external and self collisions, and the absence of a dynamic model.}
The future work consists in testing other mechanisms to understand if we can decrease the number of $\dof$s without interfering on the mesh performances. For example, we can pass from spherical to universal joints adding further constraints in the kinematic model.
Later we intend  to perform simulations with a dynamical model to estimate motor torques: they can be used in the presented genetic algorithm to further improve the actuation position scheme. Finally, we intend to actuate the cover and perform real experiments.
\change{
As for the \emph{soft external membrane}, the design and the integration is planned to start as soon as we produced a fully working rigid skeleton.
}

%% file: 8-appendix.tex
\section*{Appendix} \hypertarget{labelAppendix}{}

\begin{proofCorollary}

The matrix $\nullSpaceMatrix[\VbAndNodesRelOmega] \in \R{\dof \times 6 + 3\numberSphericalJoints}$ contains the basis of the space of the feasible movements, it is full rank and it has more rows than columns. Assume that the indexes of the linear independent rows are known and the non singular matrix $\nullSpaceMatrix[act]$ has been constructed.
Using a permutation matrix $\permm{\nullSpaceMatrix}$ to change the order of the rows, we have
\begin{equation} \label{eq:Znu partitioned}
    \nullSpaceMatrix[\VbAndNodesRelOmega] = \permm{\nullSpaceMatrix} \bm{\nullSpaceMatrix[\text{act}] \\ \nullSpaceMatrix[0]} .
\end{equation}
Given the generic solution $\VbAndNodesRelOmega = \nullSpaceMatrix[\VbAndNodesRelOmega] \boldsymbol{\xi}$, doing a change of variable $\boldsymbol{\xi} = \nullSpaceMatrix[\text{act}]^{-1} \boldsymbol{\psi}$ and substituting \eqref{eq:Znu partitioned},we obtain
\begin{equation} \label{eq:Psi}
    \VbAndNodesRelOmega = \permm{\nullSpaceMatrix} \bm{ \eye{\dof} \\ \nullSpaceMatrix[0] \nullSpaceMatrix[\text{act}]^{-1} } \boldsymbol{\psi} =
    \permm{\nullSpaceMatrix} \bm{ \boldsymbol{\psi} \\ \nullSpaceMatrix[0] \nullSpaceMatrix[\text{act}]^{-1}  \boldsymbol{\psi} } .
\end{equation}
Imposing the values of $\boldsymbol{\psi}$ is equivalent to control the relative angular velocity of some joints that are also decoupled. If we imagine to actuate those joints, we can then fully control our mesh, in a range near to this configuration, with a number of motors equal to $\dof$s.
The term $\boldsymbol{\psi}$ is called $\nodesRelOmegaAct$ in the remainder of this article (see \cref{def:nodesRelOmegaAct}).

\end{proofCorollary}

\begin{proofLemma}
Given \eqref{eq:Psi}, substituting $\nodesRelOmegaAct$ instead of $\boldsymbol{\psi}$, we can prove \eqref{eq:fromInputToRelC}
\begin{equation} \label{eq:proofNU}
    \VbAndNodesRelOmega = \nullSpaceMatrix[\VbAndNodesRelOmega] \nullSpaceMatrix[\text{act}]^{-1} \boldsymbol{\psi} = \nullSpaceMatrix[\VbAndNodesRelOmega] \nullSpaceMatrix[\text{act}]^{-1} \nodesRelOmegaAct .
\end{equation}

To proof \eqref{eq:fromInputToAbsC}, we need to recall $\Jc \nodesAbsVel = \zero{}$ and its generic solution $\nodesAbsVel = \nullSpaceMatrix[\nodesAbsVel] \boldsymbol{\xi}$.
Substituting the latter into \eqref{eq:convmV2Nu}, we obtain
\begin{equation} \label{eq:lemmaBEq1}
    \VbAndNodesRelOmega = \convm{\VbAndNodesRelOmega}{\nodesAbsVel} \nodesAbsVel = \convm{\VbAndNodesRelOmega}{\nodesAbsVel} \nullSpaceMatrix[\nodesAbsVel] \boldsymbol{\xi}.
\end{equation}
Given \eqref{eq:convmZv2Znu}, we can rewrite \eqref{eq:lemmaBEq1} as $\VbAndNodesRelOmega = \nullSpaceMatrix[\VbAndNodesRelOmega] \boldsymbol{\xi}$. Both the  absolute and relative cover velocities $\VbAndNodesRelOmega$ and $\nodesAbsVel$ are written as function of the same vector $\boldsymbol{\xi}$. Consequently, as done in \eqref{eq:Psi} and \eqref{eq:proofNU}, doing a change of variable $\boldsymbol{\xi} = \nullSpaceMatrix[\text{act}]^{-1} \nodesRelOmegaAct$ in the equation $\nodesAbsVel = \nullSpaceMatrix[\nodesAbsVel] \boldsymbol{\xi}$, we prove \eqref{eq:fromInputToAbsC}
\begin{equation}
    \nodesAbsVel = \nullSpaceMatrix[\nodesAbsVel] \nullSpaceMatrix[\text{act}]^{-1} \nodesRelOmegaAct.
\end{equation}

\end{proofLemma}

%% file: root.bbl
\begin{thebibliography}{10}
\providecommand{\url}[1]{#1}
\csname url@samestyle\endcsname
\providecommand{\newblock}{\relax}
\providecommand{\bibinfo}[2]{#2}
\providecommand{\BIBentrySTDinterwordspacing}{\spaceskip=0pt\relax}
\providecommand{\BIBentryALTinterwordstretchfactor}{4}
\providecommand{\BIBentryALTinterwordspacing}{\spaceskip=\fontdimen2\font plus
\BIBentryALTinterwordstretchfactor\fontdimen3\font minus
  \fontdimen4\font\relax}
\providecommand{\BIBforeignlanguage}[2]{{%
\expandafter\ifx\csname l@#1\endcsname\relax
\typeout{** WARNING: IEEEtran.bst: No hyphenation pattern has been}%
\typeout{** loaded for the language `#1'. Using the pattern for}%
\typeout{** the default language instead.}%
\else
\language=\csname l@#1\endcsname
\fi
#2}}
\providecommand{\BIBdecl}{\relax}
\BIBdecl

\bibitem{modelingFlyingBirdChap5}
\BIBentryALTinterwordspacing
C.~Pennycuick, ``Chapter 5 the feathered wings of birds,'' in \emph{Modelling
  the Flying Bird}, ser. Theoretical Ecology Series.\hskip 1em plus 0.5em minus
  0.4em\relax Academic Press, 2008, vol.~5, pp. 105 -- 134. [Online].
  Available:
  \url{http://www.sciencedirect.com/science/article/pii/S1875306X08000051}
\BIBentrySTDinterwordspacing

\bibitem{flyingSnake}
\BIBentryALTinterwordspacing
D.~Holden, J.~J. Socha, N.~D. Cardwell, and P.~P. Vlachos, ``Aerodynamics of
  the flying snake chrysopelea paradisi: how a bluff body cross-sectional shape
  contributes to gliding performance,'' \emph{Journal of Experimental Biology},
  vol. 217, no.~3, pp. 382--394, 2014. [Online]. Available:
  \url{https://jeb.biologists.org/content/217/3/382}
\BIBentrySTDinterwordspacing

\bibitem{morphingFrog}
J.~M. Guayasamin, T.~Krynak, K.~Krynak, J.~Culebras, and C.~R. Hutter,
  ``Phenotypic plasticity raises questions for taxonomically important traits:
  a remarkable new andean rainfrog (pristimantis) with the ability to change
  skin texture,'' \emph{Zoological Journal of the Linnean Society}, vol. 173,
  no.~4, pp. 913--928, 2015.

\bibitem{flyingFrog}
\BIBentryALTinterwordspacing
S.~B. Emerson and M.~A.~R. Koehl, ``The interaction of behavioral and
  morphological change in the evolution of a novel locomotor type: “flying”
  frogs,'' \emph{Evolution}, vol.~44, no.~8, pp. 1931--1946, 1990. [Online].
  Available:
  \url{https://onlinelibrary.wiley.com/doi/abs/10.1111/j.1558-5646.1990.tb04300.x}
\BIBentrySTDinterwordspacing

\bibitem{salamander}
\BIBentryALTinterwordspacing
M.~García-París and S.~M. Deban, ``A novel antipredator mechanism in
  salamanders: Rolling escape in hydromantes platycephalus,'' \emph{Journal of
  Herpetology}, vol.~29, no.~1, pp. 149--151, 1995. [Online]. Available:
  \url{http://www.jstor.org/stable/1565105}
\BIBentrySTDinterwordspacing

\bibitem{adaptiveMorphologyFloreano2016}
S.~Mintchev and D.~Floreano, ``Adaptive morphology: A design principle for
  multimodal and multifunctional robots,'' \emph{IEEE Robotics Automation
  Magazine}, vol.~23, no.~3, pp. 42--54, 2016.

\bibitem{morphingQuadrotorScaramuzza2019}
D.~{Falanga}, K.~{Kleber}, S.~{Mintchev}, D.~{Floreano}, and D.~{Scaramuzza},
  ``The foldable drone: A morphing quadrotor that can squeeze and fly,''
  \emph{IEEE Robotics and Automation Letters}, vol.~4, no.~2, pp. 209--216,
  April 2019.

\bibitem{roboticSkinBottiglio2019}
D.~S. {Shah}, M.~C. {Yuen}, L.~G. {Tilton}, E.~J. {Yang}, and
  R.~{Kramer-Bottiglio}, ``Morphing robots using robotic skins that sculpt
  clay,'' \emph{IEEE Robotics and Automation Letters}, vol.~4, no.~2, pp.
  2204--2211, 2019.

\bibitem{shah2020shape}
\BIBentryALTinterwordspacing
D.~Shah, B.~Yang, S.~Kriegman, M.~Levin, J.~Bongard, and R.~Kramer-Bottiglio,
  ``Shape changing robots: Bioinspiration, simulation, and physical
  realization,'' \emph{Advanced Materials}, p. 2002882, 2020. [Online].
  Available:
  \url{https://onlinelibrary.wiley.com/doi/abs/10.1002/adma.202002882}
\BIBentrySTDinterwordspacing

\bibitem{natale2017icub}
L.~Natale, C.~Bartolozzi, D.~Pucci, A.~Wykowska, and G.~Metta, ``icub: The
  not-yet-finished story of building a robot child,'' \emph{Science Robotics},
  vol.~2, no.~13, 2017.

\bibitem{atlasArticle}
E.~{Guizzo}, ``By leaps and bounds: An exclusive look at how boston dynamics is
  redefining robot agility,'' \emph{IEEE Spectrum}, vol.~56, no.~12, pp.
  34--39, 2019.

\bibitem{asimo2019}
\BIBentryALTinterwordspacing
S.~Shigemi, \emph{ASIMO and Humanoid Robot Research at Honda}.\hskip 1em plus
  0.5em minus 0.4em\relax Dordrecht: Springer Netherlands, 2019, pp. 55--90.
  [Online]. Available: \url{https://doi.org/10.1007/978-94-007-6046-2_9}
\BIBentrySTDinterwordspacing

\bibitem{surveyHumanoidRobot2019}
S.~Saeedvand, M.~Jafari, H.~S. Aghdasi, and J.~Baltes, ``A comprehensive survey
  on humanoid robot development,'' \emph{The Knowledge Engineering Review},
  vol.~34, p. e20, 2019.

\bibitem{ironcubRal}
D.~Pucci, S.~Traversaro, and F.~Nori, ``Momentum control of an underactuated
  flying humanoid robot,'' \emph{IEEE Robotics and Automation Letters}, vol.~3,
  no.~1, pp. 195--202, 2018.

\bibitem{ironcubHumanoids}
G.~Nava, L.~Fiorio, S.~Traversaro, and D.~Pucci, ``Position and attitude
  control of an underactuated flying humanoid robot,'' in \emph{2018 IEEE-RAS
  18th International Conference on Humanoid Robots (Humanoids)}, 2018, pp.
  1--9.

\bibitem{ironcubHosam}
H.~A.~O. Mohamed, G.~Nava, G.~L’Erario, S.~Traversaro, F.~Bergonti,
  L.~Fiorio, P.~R. Vanteddu, F.~Braghin, and D.~Pucci, ``Momentum-based
  extended kalman filter for thrust estimation on flying multibody robots,''
  \emph{IEEE Robotics and Automation Letters}, vol.~7, no.~1, pp. 526--533,
  2022.

\bibitem{humanoidRobotSoftSensors}
\BIBentryALTinterwordspacing
R.~Tajima, S.~Kagami, M.~Inaba, and H.~Inoue, ``Development of soft and
  distributed tactile sensors and the application to a humanoid robot,''
  \emph{Advanced Robotics}, vol.~16, no.~4, pp. 381--397, 2002. [Online].
  Available: \url{https://doi.org/10.1163/15685530260174548}
\BIBentrySTDinterwordspacing

\bibitem{hakozaki1999telemetric}
M.~Hakozaki, K.~Nakamura, and H.~Shinoda, ``Telemetric artificial skin for soft
  robot,'' in \emph{Proceedings of TRANSDUCERS}, vol.~99.\hskip 1em plus 0.5em
  minus 0.4em\relax Citeseer, 1999, pp. 844--847.

\bibitem{humanoidRobotShockAbs}
M.~Hayashi, T.~Yoshikai, and M.~Inaba, ``Development of a humanoid with
  distributed multi-axis deformation sense with full-body soft plastic foam
  cover as flesh of a robot,'' \emph{Sensors: Focus on Tactile Force and Stress
  Sensors}, pp. 319--324, 2008.

\bibitem{morphingRoboticArm2017}
A.~{Stilli}, L.~{Grattarola}, H.~{Feldmann}, H.~A. {Wurdemann}, and
  K.~{Althoefer}, ``Variable stiffness link (vsl): Toward inherently safe
  robotic manipulators,'' in \emph{2017 IEEE International Conference on
  Robotics and Automation (ICRA)}, May 2017, pp. 4971--4976.

\bibitem{origamiLang1996}
\BIBentryALTinterwordspacing
R.~J. Lang, ``A computational algorithm for origami design,'' in
  \emph{Proceedings of the Twelfth Annual Symposium on Computational Geometry},
  ser. SCG '96.\hskip 1em plus 0.5em minus 0.4em\relax New York, NY, USA:
  Association for Computing Machinery, 1996, p. 98–105. [Online]. Available:
  \url{https://doi.org/10.1145/237218.237249}
\BIBentrySTDinterwordspacing

\bibitem{rus2018origamiRobotReview}
D.~Rus and M.~T. Tolley, ``Design, fabrication and control of origami robots,''
  \emph{Nature Reviews Materials}, vol.~3, no.~6, pp. 101--112, 06 2018.

\bibitem{metaOrigamiPneumoActuated2016}
J.~T.~B. Overvelde, T.~A.~d. Jong, Y.~Shevchenko, S.~A. Becerra, G.~M.
  Whitesides, J.~C. Weaver, C.~Hoberman, and K.~Bertoldi, ``{A
  three-dimensional actuated origami-inspired transformable metamaterial with
  multiple degrees of freedom},'' \emph{Nature Communications}, vol.~7, no.~1,
  p. 10929, 2016.

\bibitem{softMorphingOrigami2019}
\BIBentryALTinterwordspacing
W.~Kim, J.~Byun, J.-K. Kim, W.-Y. Choi, K.~Jakobsen, J.~Jakobsen, D.-Y. Lee,
  and K.-J. Cho, ``Bioinspired dual-morphing stretchable origami,''
  \emph{Science Robotics}, vol.~4, no.~36, p. eaay3493, 2019. [Online].
  Available: \url{https://www.science.org/doi/abs/10.1126/scirobotics.aay3493}
\BIBentrySTDinterwordspacing

\bibitem{origamiBot2014}
E.~Vander~Hoff, D.~Jeong, and K.~Lee, ``Origamibot-i: A thread-actuated origami
  robot for manipulation and locomotion,'' in \emph{2014 IEEE/RSJ International
  Conference on Intelligent Robots and Systems}, 2014, pp. 1421--1426.

\bibitem{origamiActuatedRuss2014}
\BIBentryALTinterwordspacing
B.~An and D.~Rus, ``Designing and programming self-folding sheets,''
  \emph{Robotics and Autonomous Systems}, vol.~62, no.~7, pp. 976 -- 1001,
  2014, reconfigurable Modular Robotics. [Online]. Available:
  \url{http://www.sciencedirect.com/science/article/pii/S0921889013001528}
\BIBentrySTDinterwordspacing

\bibitem{walkingOrigamiRus2015}
S.~Miyashita, S.~Guitron, M.~Ludersdorfer, C.~R. Sung, and D.~Rus, ``An
  untethered miniature origami robot that self-folds, walks, swims, and
  degrades,'' in \emph{2015 IEEE International Conference on Robotics and
  Automation (ICRA)}, 2015, pp. 1490--1496.

\bibitem{recSurfOriPixelEPFL2020}
M.~{Salerno}, J.~{Paik}, and S.~{Mintchev}, ``Ori-pixel, a multi-dofs origami
  pixel for modular reconfigurable surfaces,'' \emph{IEEE Robotics and
  Automation Letters}, vol.~5, no.~4, pp. 6988--6995, 2020.

\bibitem{recSurfMouldingPatent}
M.~K. Kristensen and C.~R. Jepsen, ``Flexible mat for providing a dynamically
  reconfigurable double-curved moulding surface in a mould,'' Oct.~27 2015, uS
  Patent 9,168,678.

\bibitem{recSurfESA}
\BIBentryALTinterwordspacing
{ARTES ESA}, ``Tas developing a reconfigurable antenna,'' 2016. [Online].
  Available:
  \url{https://artes.esa.int/news/tas-developing-reconfigurable-antenna}
\BIBentrySTDinterwordspacing

\bibitem{recSurfFesto}
\BIBentryALTinterwordspacing
{Festo Corporate}, ``Wavehandling,'' 2014. [Online]. Available:
  \url{https://www.festo.com/group/en/cms/10225.htm}
\BIBentrySTDinterwordspacing

\bibitem{recSurfInForm2013}
\BIBentryALTinterwordspacing
S.~Follmer, D.~Leithinger, A.~Olwal, A.~Hogge, and H.~Ishii, ``Inform: Dynamic
  physical affordances and constraints through shape and object actuation,'' in
  \emph{Proceedings of the 26th Annual ACM Symposium on User Interface Software
  and Technology}.\hskip 1em plus 0.5em minus 0.4em\relax Association for
  Computing Machinery, 2013, p. 417–426. [Online]. Available:
  \url{https://doi.org/10.1145/2501988.2502032}
\BIBentrySTDinterwordspacing

\bibitem{softRobotReview2015}
D.~Rus and M.~T. Tolley, ``{Design, fabrication and control of soft robots},''
  \emph{Nature}, vol. 521, no. 7553, pp. 467--475, 2015.

\bibitem{softRobotReview2017}
C.~Lee, M.~Kim, Y.~J. Kim, N.~Hong, S.~Ryu, H.~J. Kim, and S.~Kim, ``Soft robot
  review,'' \emph{International Journal of Control, Automation and Systems},
  vol.~15, no.~1, pp. 3--15, 2017.

\bibitem{softRobotReview2018}
T.~Wallin, J.~Pikul, and R.~Shepherd, ``3d printing of soft robotic systems,''
  \emph{Nature Reviews Materials}, vol.~3, no.~6, pp. 84--100, 2018.

\bibitem{activeTextile2017}
Y.~Funabora, ``Prototype of a fabric actuator with multiple thin artificial
  muscles for wearable assistive devices,'' in \emph{2017 IEEE/SICE
  International Symposium on System Integration (SII)}, 2017, pp. 356--361.

\bibitem{activeTextile2019}
T.~Hiramitsu, K.~Suzumori, H.~Nabae, and G.~Endo, ``Experimental evaluation of
  textile mechanisms made of artificial muscles,'' in \emph{2019 2nd IEEE
  International Conference on Soft Robotics (RoboSoft)}, 2019, pp. 1--6.

\bibitem{softRoboticPad2017}
Y.~Sun, J.~Guo, T.~M. Miller-Jackson, X.~Liang, M.~H. Ang, and R.~C.~H. Yeow,
  ``Design and fabrication of a shape-morphing soft pneumatic actuator: Soft
  robotic pad,'' in \emph{2017 IEEE/RSJ International Conference on Intelligent
  Robots and Systems (IROS)}, 2017, pp. 6214--6220.

\bibitem{bucklingSoftGel2012}
\BIBentryALTinterwordspacing
J.~Kim, J.~A. Hanna, M.~Byun, C.~D. Santangelo, and R.~C. Hayward, ``Designing
  responsive buckled surfaces by halftone gel lithography,'' \emph{Science},
  vol. 335, no. 6073, pp. 1201--1205, 2012. [Online]. Available:
  \url{https://www.science.org/doi/abs/10.1126/science.1215309}
\BIBentrySTDinterwordspacing

\bibitem{morphingPasta2021}
\BIBentryALTinterwordspacing
Y.~Tao, Y.-C. Lee, H.~Liu, X.~Zhang, J.~Cui, C.~Mondoa, M.~Babaei,
  J.~Santillan, G.~Wang, D.~Luo, D.~Liu, H.~Yang, Y.~Do, L.~Sun, W.~Wang,
  T.~Zhang, and L.~Yao, ``Morphing pasta and beyond,'' \emph{Science Advances},
  vol.~7, no.~19, p. eabf4098, 2021. [Online]. Available:
  \url{https://www.science.org/doi/abs/10.1126/sciadv.abf4098}
\BIBentrySTDinterwordspacing

\bibitem{aeroMorph2016}
\BIBentryALTinterwordspacing
J.~Ou, M.~Skouras, N.~Vlavianos, F.~Heibeck, C.-Y. Cheng, J.~Peters, and
  H.~Ishii, ``Aeromorph - heat-sealing inflatable shape-change materials for
  interaction design,'' in \emph{Proceedings of the 29th Annual Symposium on
  User Interface Software and Technology}, ser. UIST '16.\hskip 1em plus 0.5em
  minus 0.4em\relax New York, NY, USA: Association for Computing Machinery,
  2016, p. 121–132. [Online]. Available:
  \url{https://doi.org/10.1145/2984511.2984520}
\BIBentrySTDinterwordspacing

\bibitem{camouflageRobot2013}
\BIBentryALTinterwordspacing
S.~Daynes, A.~Grisdale, A.~Seddon, and R.~Trask, ``Morphing structures using
  soft polymers for active deployment,'' \emph{Smart Materials and Structures},
  vol.~23, no.~1, p. 012001, dec 2013. [Online]. Available:
  \url{https://doi.org/10.1088/0964-1726/23/1/012001}
\BIBentrySTDinterwordspacing

\bibitem{camouflageRobotScience2017}
\BIBentryALTinterwordspacing
J.~H. Pikul, S.~Li, H.~Bai, R.~T. Hanlon, I.~Cohen, and R.~F. Shepherd,
  ``Stretchable surfaces with programmable 3d texture morphing for synthetic
  camouflaging skins,'' \emph{Science}, vol. 358, no. 6360, pp. 210--214, 2017.
  [Online]. Available:
  \url{https://www.science.org/doi/abs/10.1126/science.aan5627}
\BIBentrySTDinterwordspacing

\bibitem{pneUISoftRobot2013}
\BIBentryALTinterwordspacing
L.~Yao, R.~Niiyama, J.~Ou, S.~Follmer, C.~Della~Silva, and H.~Ishii, ``Pneui:
  Pneumatically actuated soft composite materials for shape changing
  interfaces,'' in \emph{Proceedings of the 26th Annual ACM Symposium on User
  Interface Software and Technology}, ser. UIST '13.\hskip 1em plus 0.5em minus
  0.4em\relax New York, NY, USA: Association for Computing Machinery, 2013, p.
  13–22. [Online]. Available: \url{https://doi.org/10.1145/2501988.2502037}
\BIBentrySTDinterwordspacing

\bibitem{underwaterWalkingRobot2019}
M.~Ishida, D.~Drotman, B.~Shih, M.~Hermes, M.~Luhar, and M.~T. Tolley,
  ``Morphing structure for changing hydrodynamic characteristics of a soft
  underwater walking robot,'' \emph{IEEE Robotics and Automation Letters},
  vol.~4, no.~4, pp. 4163--4169, 2019.

\bibitem{softElastometerDaraio2021}
\BIBentryALTinterwordspacing
K.~Liu, F.~Hacker, and C.~Daraio, ``Robotic surfaces with reversible,
  spatiotemporal control for shape morphing and object manipulation,''
  \emph{Science Robotics}, vol.~6, no.~53, p. eabf5116, 2021. [Online].
  Available: \url{https://www.science.org/doi/abs/10.1126/scirobotics.abf5116}
\BIBentrySTDinterwordspacing

\bibitem{4Dprinting2019}
\BIBentryALTinterwordspacing
X.~Kuang, D.~J. Roach, J.~Wu, C.~M. Hamel, Z.~Ding, T.~Wang, M.~L. Dunn, and
  H.~J. Qi, ``Advances in 4d printing: Materials and applications,''
  \emph{Advanced Functional Materials}, vol.~29, no.~2, p. 1805290, 2019.
  [Online]. Available:
  \url{https://onlinelibrary.wiley.com/doi/abs/10.1002/adfm.201805290}
\BIBentrySTDinterwordspacing

\bibitem{4DprintingWater2014}
D.~Raviv, W.~Zhao, C.~McKnelly, A.~Papadopoulou, A.~Kadambi, B.~Shi, S.~Hirsch,
  D.~Dikovsky, M.~Zyracki, C.~Olguin, R.~Raskar, and S.~Tibbits, ``{Active
  Printed Materials for Complex Self-Evolving Deformations},'' \emph{Scientific
  Reports}, vol.~4, no.~1, p. 7422, 2014.

\bibitem{4DprintingTemperature2019}
\BIBentryALTinterwordspacing
J.~W. Boley, W.~M. van Rees, C.~Lissandrello, M.~N. Horenstein, R.~L. Truby,
  A.~Kotikian, J.~A. Lewis, and L.~Mahadevan, ``Shape-shifting structured
  lattices via multimaterial 4d printing,'' \emph{Proceedings of the National
  Academy of Sciences}, vol. 116, no.~42, pp. 20\,856--20\,862, 2019. [Online].
  Available: \url{https://www.pnas.org/content/116/42/20856}
\BIBentrySTDinterwordspacing

\bibitem{magneticSoftRobot2018}
Y.~Kim, H.~Yuk, R.~Zhao, S.~A. Chester, and X.~Zhao, ``{Printing ferromagnetic
  domains for untethered fast-transforming soft materials},'' \emph{Nature},
  vol. 558, no. 7709, pp. 274--279, 2018.

\bibitem{morphingWingReview2010}
\BIBentryALTinterwordspacing
A.~Sofla, S.~Meguid, K.~Tan, and W.~Yeo, ``Shape morphing of aircraft wing:
  Status and challenges,'' \emph{Materials \& Design}, vol.~31, no.~3, pp.
  1284--1292, 2010. [Online]. Available:
  \url{https://www.sciencedirect.com/science/article/pii/S0261306909004968}
\BIBentrySTDinterwordspacing

\bibitem{morphingWingReview2011}
\BIBentryALTinterwordspacing
S.~Barbarino, O.~Bilgen, R.~M. Ajaj, M.~I. Friswell, and D.~J. Inman, ``A
  review of morphing aircraft,'' \emph{Journal of Intelligent Material Systems
  and Structures}, vol.~22, no.~9, pp. 823--877, 2011. [Online]. Available:
  \url{https://doi.org/10.1177/1045389X11414084}
\BIBentrySTDinterwordspacing

\bibitem{morphingSkins2008}
C.~Thill, J.~Etches, I.~Bond, K.~Potter, and P.~Weaver, ``Morphing skins,''
  \emph{The Aeronautical Journal (1968)}, vol. 112, no. 1129, p. 117–139,
  2008.

\bibitem{planformMorphingWing2004}
\BIBentryALTinterwordspacing
D.~Neal, M.~Good, C.~Johnston, H.~Robertshaw, W.~Mason, and D.~Inman,
  \emph{Design and Wind-Tunnel Analysis of a Fully Adaptive Aircraft
  Configuration}.\hskip 1em plus 0.5em minus 0.4em\relax AIAA, 2004. [Online].
  Available: \url{https://arc.aiaa.org/doi/abs/10.2514/6.2004-1727}
\BIBentrySTDinterwordspacing

\bibitem{planformMorphingWing2007}
\BIBentryALTinterwordspacing
J.~Flanagan, R.~Strutzenberg, R.~Myers, and J.~Rodrian, \emph{Development and
  Flight Testing of a Morphing Aircraft, the NextGen MFX-1}.\hskip 1em plus
  0.5em minus 0.4em\relax AIAA, 2007. [Online]. Available:
  \url{https://arc.aiaa.org/doi/abs/10.2514/6.2007-1707}
\BIBentrySTDinterwordspacing

\bibitem{optimalActuatorMorphingWing2006}
\BIBentryALTinterwordspacing
J.~J. Joo, B.~Sanders, T.~Johnson, and M.~I. Frecker, ``{Optimal actuator
  location within a morphing wing scissor mechanism configuration},'' in
  \emph{Smart Structures and Materials 2006: Modeling, Signal Processing, and
  Control}, D.~K. Lindner, Ed., vol. 6166, International Society for Optics and
  Photonics.\hskip 1em plus 0.5em minus 0.4em\relax SPIE, 2006, pp. 24 -- 35.
  [Online]. Available: \url{https://doi.org/10.1117/12.658830}
\BIBentrySTDinterwordspacing

\bibitem{oopMorphingWing2007}
\BIBentryALTinterwordspacing
R.~Vos, R.~Barrett, R.~de~Breuker, and P.~Tiso, ``Post-buckled precompressed
  elements: a new class of control actuators for morphing wing {UAVs},''
  \emph{Smart Materials and Structures}, vol.~16, no.~3, pp. 919--926, may
  2007. [Online]. Available: \url{https://doi.org/10.1088/0964-1726/16/3/042}
\BIBentrySTDinterwordspacing

\bibitem{airfoilMorphingWing2013}
\BIBentryALTinterwordspacing
O.~Bilgen, L.~M. Butt, S.~R. Day, C.~A. Sossi, J.~P. Weaver, A.~Wolek, W.~H.
  Mason, and D.~J. Inman, ``A novel unmanned aircraft with solid-state control
  surfaces: Analysis and flight demonstration,'' \emph{Journal of Intelligent
  Material Systems and Structures}, vol.~24, no.~2, pp. 147--167, 2013.
  [Online]. Available: \url{https://doi.org/10.1177/1045389X12459592}
\BIBentrySTDinterwordspacing

\bibitem{modularRobotReview2007}
M.~{Yim}, W.~{Shen}, B.~{Salemi}, D.~{Rus}, M.~{Moll}, H.~{Lipson},
  E.~{Klavins}, and G.~S. {Chirikjian}, ``Modular self-reconfigurable robot
  systems [grand challenges of robotics],'' \emph{IEEE Robotics Automation
  Magazine}, vol.~14, no.~1, pp. 43--52, 2007.

\bibitem{modularRobotReview2015}
\BIBentryALTinterwordspacing
H.~Ahmadzadeh and E.~Masehian, ``Modular robotic systems: Methods and
  algorithms for abstraction, planning, control, and synchronization,''
  \emph{Artificial Intelligence}, vol. 223, pp. 27 -- 64, 2015. [Online].
  Available:
  \url{http://www.sciencedirect.com/science/article/pii/S0004370215000260}
\BIBentrySTDinterwordspacing

\bibitem{morpho2008}
C.-H. Yu, K.~Haller, D.~Ingber, and R.~Nagpal, ``Morpho: A self-deformable
  modular robot inspired by cellular structure,'' in \emph{2008 IEEE/RSJ
  International Conference on Intelligent Robots and Systems}, 2008, pp.
  3571--3578.

\bibitem{mori2017}
C.~H. {Belke} and J.~{Paik}, ``Mori: A modular origami robot,'' \emph{IEEE/ASME
  Transactions on Mechatronics}, vol.~22, no.~5, pp. 2153--2164, 2017.

\bibitem{moriAutomaticCoupling2019}
C.~H. Belke and J.~Paik, ``Automatic couplings with mechanical overload
  protection for modular robots,'' \emph{IEEE/ASME Transactions on
  Mechatronics}, vol.~24, no.~3, pp. 1420--1426, 2019.

\bibitem{moriReconfigurationStrategy2019}
\BIBentryALTinterwordspacing
M.~Yao, C.~H. Belke, H.~Cui, and J.~Paik, ``A reconfiguration strategy for
  modular robots using origami folding,'' \emph{The International Journal of
  Robotics Research}, vol.~38, no.~1, pp. 73--89, 2019. [Online]. Available:
  \url{https://doi.org/10.1177/0278364918815757}
\BIBentrySTDinterwordspacing

\bibitem{moriOptimalDistributionActiveModules}
\BIBentryALTinterwordspacing
M.~Yao, X.~Xiao, C.~H. Belke, H.~Cui, and J.~Paik, ``{Optimal Distribution of
  Active Modules in Reconfiguration Planning of Modular Robots},''
  \emph{Journal of Mechanisms and Robotics}, vol.~11, no.~1, 12 2018, 011017.
  [Online]. Available: \url{https://doi.org/10.1115/1.4041972}
\BIBentrySTDinterwordspacing

\bibitem{goldberg1988genetic}
D.~E. Goldberg and J.~H. Holland, \emph{Genetic algorithms and machine
  learning}.\hskip 1em plus 0.5em minus 0.4em\relax Kluwer Academic
  Publishers-Plenum Publishers, 1988.

\bibitem{mitchell1998introduction}
M.~Mitchell, \emph{An introduction to genetic algorithms}.\hskip 1em plus 0.5em
  minus 0.4em\relax MIT press, 1998.

\bibitem{selectionGA}
J.~Z. Jinghui~{Zhong}, Xiaomin~{Hu} and M.~{Gu}, ``Comparison of performance
  between different selection strategies on simple genetic algorithms,'' in
  \emph{International Conference on Computational Intelligence for Modelling,
  Control and Automation and International Conference on Intelligent Agents,
  Web Technologies and Internet Commerce (CIMCA-IAWTIC'06)}, vol.~2, 2005, pp.
  1115--1121.

\bibitem{crossoverGA}
W.~M. Spears and V.~Anand, ``A study of crossover operators in genetic
  programming,'' in \emph{Methodologies for Intelligent Systems}, Z.~W. Ras and
  M.~Zemankova, Eds.\hskip 1em plus 0.5em minus 0.4em\relax Berlin, Heidelberg:
  Springer Berlin Heidelberg, 1991, pp. 409--418.

\bibitem{roboticsHandbook}
B.~Siciliano and O.~Khatib, \emph{Springer handbook of robotics}.\hskip 1em
  plus 0.5em minus 0.4em\relax Springer, 2016.

\bibitem{PUCCI201572}
\BIBentryALTinterwordspacing
D.~Pucci, T.~Hamel, P.~Morin, and C.~Samson, ``Nonlinear feedback control of
  axisymmetric aerial vehicles,'' \emph{Automatica}, vol.~53, pp. 72--78, 2015.
  [Online]. Available:
  \url{http://www.sciencedirect.com/science/article/pii/S0005109814006104}
\BIBentrySTDinterwordspacing

\bibitem{dafarra2020}
S.~Dafarra, G.~Romualdi, G.~Metta, and D.~Pucci, ``Whole-body walking
  generation using contact parametrization: A non-linear trajectory
  optimization approach,'' in \emph{2020 IEEE International Conference on
  Robotics and Automation (ICRA)}, 2020, pp. 1511--1517.

\bibitem{osqp}
\BIBentryALTinterwordspacing
B.~Stellato, G.~Banjac, P.~Goulart, A.~Bemporad, and S.~Boyd, ``{OSQP}: an
  operator splitting solver for quadratic programs,'' \emph{Mathematical
  Programming Computation}, 2020. [Online]. Available:
  \url{https://doi.org/10.1007/s12532-020-00179-2}
\BIBentrySTDinterwordspacing

\bibitem{casadi}
J.~A.~E. Andersson, J.~Gillis, G.~Horn, J.~B. Rawlings, and M.~Diehl,
  ``{CasADi} -- {A} software framework for nonlinear optimization and optimal
  control,'' \emph{Mathematical Programming Computation}, vol.~11, no.~1, pp.
  1--36, 2019.

\bibitem{Baumgarte}
S.~{Gros}, M.~{Zanon}, and M.~{Diehl}, ``Baumgarte stabilisation over the so(3)
  rotation group for control,'' in \emph{2015 54th IEEE Conference on Decision
  and Control (CDC)}, Dec 2015, pp. 620--625.

\bibitem{choosingParametersGA}
\BIBentryALTinterwordspacing
A.~Hassanat, K.~Almohammadi, E.~Alkafaween, E.~Abunawas, A.~Hammouri, and
  V.~B.~S. Prasath, ``Choosing mutation and crossover ratios for genetic
  algorithms—a review with a new dynamic approach,'' \emph{Information},
  vol.~10, no.~12, p. 390, Dec 2019. [Online]. Available:
  \url{http://dx.doi.org/10.3390/info10120390}
\BIBentrySTDinterwordspacing

\end{thebibliography}
